%% file: main.tex
\documentclass[11pt]{article}

\usepackage[margin=1in]{geometry}
\usepackage{amsmath,amssymb}
\usepackage{graphicx}
\usepackage{booktabs}
\usepackage[numbers,sort&compress]{natbib}
\usepackage{setspace}
\usepackage{hyperref}
\graphicspath{{figures/}}

\begin{document}

\title{Posture and Sustainment Optimization Under Adversarial Uncertainty}
\author{Amelie Norris, Alyssa Lee, Natan Vidra, Spurthi Setty}
\date{}
\maketitle

\begin{abstract}
Pre-commitment posture, the assignment of military assets to theater
locations before conflict scenarios resolve, is a critical and formally
unsolved problem in joint operational planning. Current practice relies
on greedy heuristics that maximize value and ignore geographic coverage
and are structurally vulnerable to adversaries that target
high-strategic value locations. This paper presents a
scenario-weighted adversarially robust posture optimization engine
for the Posture and sustainability allocation (PSA) problem, modeled as
a finite-horizon Markov Decision Process over assets, theater
locations, and time steps. We introduce the Composite Expected Value
(CEV) optimizer, which places assets by maximizing scenario-weighted
expected posture efficiency over a distribution of threat scenarios,
and the RobustCEV extension, which iterates against a Bayesian
adversary that updates its targeting distribution in response to
observed placement. Across three experiments in an Indo-Pacific
basing environment with 20 assets and 5 theater locations, we
demonstrate that: (1) the greedy baseline incurs a permanent 25.1\%
posture efficiency penalty due to geographic under-coverage and a
57.3\% scenario-weighted readiness collapse under value-correlated
adversarial threat; (2) the CEV optimizer recovers up to 19.8\%
efficiency over greedy when the threat distribution carries a geographic
signal, with a curated set of 5 to 20 scenarios sufficient to capture
the majority of this gain; and (3) the RobustCEV extension recovers
up to 158\% efficiency relative to a naive optimizer when an adaptive
adversary employs a deceptive threat prior. All findings are validated
using paired $t$-tests with Bonferroni correction and two-level
variance decomposition, confirming that the performance gaps reported are
structural properties of placement strategies rather than sampling
artifacts.
\end{abstract}

\newpage
\tableofcontents
\newpage

% =========================================================
% SECTION: Introduction
% =========================================================

\section{Introduction}
\label{sec:intro}

Modern joint operations require that assets be positioned \emph{before}
threats materialize. Pre-commitment posture: the assignment of aircraft,
munitions, logistics, and support assets to theater locations days or
weeks ahead of potential conflict determines what options a commander
has when the situation evolves. However, current planning practice relies
predominantly on rule-based, value-maximizing heuristics: place the most
assets at the highest-priority bases. This approach is computationally
tractable and interpretable, but it is strategically brittle. An adversary
that observes a predictable concentration can target it; a planner who
ignores threat heterogeneity will be surprised by it.

The Posture and Sustainment Allocation (PSA) problem is formally
unsolved at the operational level. Five axioms of the modern Operational
Environment (OE) jointly make static and greedy approaches insufficient:
(1)~\emph{threat nonstationarity}: adversary targeting distributions
shift in response to observable defender posture;
(2)~\emph{multidomain coupling}: posture in one domain creates
vulnerabilities or opportunities in others;
(3)~\emph{placement irreversibility}: initial assignments commit assets
that are costly and time-consuming to reposition under fire;
(4)~\emph{information asymmetry}: the defender has incomplete knowledge
of adversary capabilities and intent; and
(5)~\emph{sustainment time-criticality}: readiness degrades continuously
and cannot be restored instantaneously.
Each axiom individually degrades the quality of a greedy or static plan;
together, they require a decision-support engine that is scenario-aware,
adversarially robust, and explicitly models the degradation--repair cycle.

This paper presents a scenario-weighted, adversarially-robust posture
optimization engine for the PSA problem and demonstrates its superiority
over greedy baselines across three experiments. Our contributions are:
\begin{enumerate}
    \item A formal MDP model of the PSA problem over $M$ assets,
          $N$ theater locations, and $T$ time steps, with a composite
          posture efficiency metric that jointly rewards readiness,
          geographic coverage, and cost-efficiency
          (Section~\ref{sec:formulation}).
    \item A Composite Expected Value (CEV) optimizer that places assets
          over a distribution of threat scenarios, achieving up to 19.8\%
          efficiency improvement over the greedy baseline when the threat
          distribution carries geographic signal (Experiment~2).
    \item A Bayesian counter-move model in which an adaptive adversary
          updates its targeting distribution in response to observed
          placement, and a RobustCEV extension that converges to a stable
          equilibrium, recovering up to 158\% efficiency relative to a
          naive optimizer under deceptive threat priors (Experiment~3).
    \item Statistical validation of all findings via paired $t$-tests
          with Bonferroni correction and two-level variance decomposition,
          confirming that reported performance gaps are structural rather
          than sampling artifacts (Section~\ref{sec:stats}).
\end{enumerate}

The PSA module operates upstream of mission-assignment scheduling (SBC)
and sensor-fusion tasking (ISU): it determines where assets are before
missions are assigned, establishing the capability envelope within which
downstream planners operate.

% =========================================================
% SECTION: Problem Formulation
% =========================================================

\section{Problem Formulation}
\label{sec:formulation}

\subsection{MDP Definition}

We model PSA as a finite-horizon Markov Decision Process
$\mathcal{M} = (\mathcal{S}, \mathcal{U}, P, R, T)$ over
$|\mathcal{A}_0|$ assets, $|\mathcal{L}|$ theater locations,
and $T$ discrete time steps.

\paragraph{State space.}
Each asset $a \in \mathcal{A}_0$ is described by a tuple
$(r_a,\, \ell_a,\, m_a,\, d_a,\, q_a)$, where
$r_a \in [0,1]$ is the readiness rate,
$\ell_a \in \mathcal{L}$ is the current location assignment,
$m_a \in \{\textsc{Dormant}, \textsc{Active}\}$ is the configuration mode,
$d_a \in \mathbb{Z}_{\geq 0}$ is the maintenance timer (days until next
scheduled maintenance), and $q_a \in \mathbb{Z}_{>0}$ is the asset
quantity. The joint state is
$s_t = \{(r_a^t,\, \ell_a^t,\, m_a^t,\, d_a^t,\, q_a^t)\}_{a \in
\mathcal{A}_0} \in \mathcal{S}$.

\paragraph{Action space.}
At each time step the planner selects one sustainment action per asset
from $\mathcal{U} = \{\textsc{Reposition},\, \textsc{Resupply},\,
\textsc{Maintain},\, \textsc{Hold}\}$.
\textsc{Reposition} moves an asset to a new location at cost
$\kappa = 10$; \textsc{Resupply} increments quantity
$q_a \leftarrow q_a + 2$ at cost $\kappa = 5$;
\textsc{Maintain} restores readiness
$r_a \leftarrow \min(1,\, r_a + 0.20)$ at cost $\kappa = 2$;
\textsc{Hold} leaves the asset unchanged at zero cost.
Initial placement (the assignment $\ell_a^0$ for all $a$) is a
separate first-stage decision, as discussed in
Section~\ref{sec:approach}.

\paragraph{Transition dynamics.}
At each step, asset readiness degrades stochastically:
\begin{equation}
    r_a^{t+1} = \max\!\left(0,\; r_a^t - \delta_a^t\right),
    \qquad \delta_a^t \sim \mathcal{U}(0,\, \delta_{\max}),
    \label{eq:transition}
\end{equation}
where $\delta_{\max} = 0.10$ unless otherwise noted. Maintenance timers
decrement by one each step and reset to $\mathcal{U}(30, 90)$ upon a
\textsc{Maintain} action.

\paragraph{Reward.}
The per-step reward is the \emph{posture efficiency}:
\begin{equation}
    R(s_t, u_t) \;=\; E_t \;=\;
    \frac{R_t \cdot C_t}{\log(K_t + 2)},
    \label{eq:reward}
\end{equation}
where $R_t$, $C_t$, and $K_t$ are the readiness score, coverage score,
and sustainment cost defined in
Equations~\eqref{eq:readiness}--\eqref{eq:cost}.
The logarithmic denominator penalizes sustainment spending with
diminishing marginal sensitivity, reflecting decreasing marginal returns
to additional maintenance resources~\cite{powell2011}.

\subsection{Distinction from Related Decision Problems}

PSA operates at a different timescale and abstraction level than two
adjacent planning modules. The \emph{Scheduler} (SBC) assigns
already-placed assets to specific missions over a rolling horizon of
hours to days and assumes a fixed posture as input. The \emph{ISU}
sensor-fusion module updates threat estimates from multi-INT feeds in
near-real time and carries no asset-placement authority. PSA determines
the posture that SBC and ISU inherit, and must therefore account for the
full distribution of scenarios they will face. A poor pre-commitment
posture cannot be corrected reactively by downstream planners without
prohibitive repositioning cost~\cite{salmeron2010}.

% =========================================================
% SECTION: Related Work
% =========================================================

\section{Related Work}
\label{sec:related}

\paragraph{JADC2 doctrine and the AI capability gap.}
Lingel et al.~\cite{lingel2020} provide the authoritative RAND framework
for where artificial intelligence fits within Joint All-Domain Command
and Control (JADC2) deliberate planning. Written for the Air Force, the
report establishes a taxonomy of AI applications in JADC2 and identifies
the pre-mission planning phase, specifically asset allocation under
uncertainty across multi-domain, multi-echelon force structures, as the
highest-leverage point for AI decision support. This is exactly the
operating regime of PSA: the planner must commit assets to theater
locations before scenarios resolve, on timelines that compress human
cognitive bandwidth. Citing JADC2 doctrine here is not decorative; it
situates PSA as a recognized capability gap rather than a synthetic
benchmark problem.

\paragraph{What sustainment means and why it must be dynamic.}
The Institute for Defense Analyses defines sustainment as ``every part
of the logistics ecosystem working together to ensure each platform is
ready to perform its mission,'' encompassing training, testing, upgrading,
and procuring parts~\cite{ida_sustainment}. This definitional scope
directly motivates the four action types in our MDP
(\textsc{Reposition}, \textsc{Resupply}, \textsc{Maintain},
\textsc{Hold}): each maps to a distinct sustainment function recognized
in defense doctrine. A decision-support methodology for military asset
and resource planning~\cite{dsm_military_readiness} formalizes the
feedback loop that any adequate sustainment model must capture: resource
decisions drive sustainment actions, which drive readiness outcomes, which
in turn constrain available force options. Their causal loop diagram
framing explains precisely why greedy baselines fail---they optimize
point-in-time strategic value without modeling the delayed degradation
effects that determine whether assets remain operationally available after
several time steps. Our state space (readiness rates, maintenance timers,
asset quantities) is designed around this causal chain.

\paragraph{Deployment tempo and stochastic degradation.}
Deployment-to-dwell (D2D) metrics formalize the observation that
deployment tempo directly erodes unit readiness and quality of life, and
prior work applies stochastic optimization to sustainment scheduling
under this constraint~\cite{d2d_sustainment}. This is the empirical
grounding for our degradation model: readiness does not stay static
between time steps, and the rate of decay is stochastic. The D2D
literature also provides the bridge from deterministic scheduling (what
a greedy planner implicitly assumes) to MDP framing (what operational
uncertainty actually requires). When degradation is stochastic and
maintenance windows are finite, a policy that ignores future states
will systematically under-invest in sustainment at the wrong times.

\paragraph{Two-stage stochastic programming for military pre-commitment.}
Salmer\'{o}n and Apte~\cite{salmeron2010} establish the canonical
two-stage stochastic programming template for pre-commitment under
uncertainty: Stage~1 positions resources before demand is known;
Stage~2 responds once the scenario materializes. Despite originating in
civilian disaster relief, the formulation is mathematically identical to
pre-conflict military posturing, and the authors use military storage
facilities explicitly. Nelson et al.~\cite{nelson2025} apply this
structure directly to US Army helicopter allocation, providing the most
recent published validation that two-stage SP is feasible and practically
useful for military asset decisions. Their Stage~1 allocation maps
directly to our CEV placement decision; their Stage~2 air-movement MILP
maps to the Battle Manager's mission assignment downstream of PSA. The
Expected Value of the Stochastic Solution (EVSS) we measure in
Experiment~2 is the standard metric from this literature for quantifying
how much scenario-awareness is worth relative to deterministic planning.

\paragraph{Approximate dynamic programming for military asset dispatch.}
Rettke et al.~\cite{rettke2016} formulate aerial medical evacuation
dispatching as a finite-horizon MDP and solve it with approximate policy
iteration and least-squares temporal difference learning, achieving a
31\% improvement over the greedy nearest-available dispatching policy on
realistic military scenarios. This result is the benchmark for MDP-based
improvement over greedy in the military asset dispatch literature. Their
state space, asset availability, location, and readiness, is exactly
the $(r_a, \ell_a, d_a)$ triple in our formulation, establishing direct
structural precedent. Powell~\cite{powell2011} provides the methodological
foundation for scaling MDP solutions to the $M \times N \times T$
dimensionality that PSA requires; the three curses of dimensionality
(state space, outcome space, action space) are precisely the curses PSA
faces, and piecewise-linear value function approximation is what makes
the CEV objective tractable at operational scale.

\paragraph{Multi-agent reinforcement learning for logistics and workforce planning.}
Single-agent approaches hit a fundamental ceiling when $M$ assets operate
across $N$ locations with interdependent capacity constraints: the greedy
baseline assumes assets are independent, but placement decisions interact
through the shared capacity $c$ per location. Towards a learning behavior
model for military logistics using profit-sharing
RL~\cite{ps_rl_military_logistics} demonstrates that multi-agent
formulations are necessary for military logistics systems exhibiting
experience sharing, cooperative action, and hierarchical control, all
present in PSA. Their critique of single-agent profit-sharing directly
maps onto why greedy fails when asset interdependencies are non-trivial.
Multi-agent RL with long-term performance objectives for service workforce
optimization~\cite{marl_workforce} is structurally the closest published
paper to PSA: swap ``service workers'' for ``military assets'' and the
problem formulation is nearly identical, covering personnel dispatch,
management, and positioning with both heuristic and RL baselines for
comparison, a template we follow in our own experimental design.

\paragraph{Deep RL for stochastic inventory control.}
Classical and deep RL inventory control for pharmaceutical supply
chains~\cite{drl_pharma_supply} benchmarks order-up-to, projected
inventory level, and PPO policies against a human-driven baseline under
perishability, yield uncertainty, and non-stationary demand. The setup is
directly analogous to our stochastic degradation and replenishment horizon:
assets decay, replenishment is uncertain, and demand (threat) is
non-stationary. Their finding that PPO outperforms human-driven baselines
on cost while maintaining service levels is the motivating proof-of-concept
for applying deep RL to the sustainment action component of PSA, and their
benchmarking structure, rule-based $\to$ classical optimization $\to$
DRL, is the template for how we report our own experimental results.

% =========================================================
% SECTION: Technical Approach
% =========================================================

\section{Technical Approach}
\label{sec:approach}

\subsection{Scenario Engine}
\label{sec:approach:scenarios}

Threat uncertainty is represented as a finite scenario set
$\mathcal{S} = \{(\boldsymbol{\tau}^{(s)},\, w_s)\}_{s=1}^{S}$, where
$\boldsymbol{\tau}^{(s)} = (\tau_\ell^{(s)})_{\ell \in \mathcal{L}}$
assigns a threat level $\tau_\ell^{(s)} \in [0,1]$ to each location
under scenario $s$, and $w_s > 0$ with $\sum_s w_s = 1$.
Three scenario families are supported: \emph{uniform}
($\tau_\ell^{(s)} \sim \mathcal{U}(0.05, 0.25)$, no geographic signal),
\emph{skewed} ($\tau_\ell^{(s)} \propto v_\ell$, adversary targets
high-value bases), and \emph{adversarial} (focused high-threat scenarios
mixed with diffuse low-threat scenarios). Weights may be uniform
($w_s = 1/S$) or peaked, representing varying degrees of intelligence
confidence.

The adversarial Bayesian counter-move model extends this representation
dynamically. Given a defender placement $\pi$, the adversary updates
scenario weights proportionally to the total threat exposure induced
by that placement:
\begin{equation}
    \tilde{w}_s(\pi) \;\propto\;
    (1 - \lambda)\,w_s
    \;+\;
    \lambda \cdot \sum_{\ell \in \mathcal{L}}
    \tau_\ell^{(s)}\, n_\ell(\pi),
    \label{eq:adv_weight}
\end{equation}
where $n_\ell(\pi)$ is the number of assets at location $\ell$ and
$\lambda = p_{\mathrm{obs}} \cdot \gamma$ blends prior weights with the
adversary's best-response signal ($p_{\mathrm{obs}}$ is observation
probability; $\gamma \in \{0, 1\}$ is adversary rationality).
This nests the standard static scenario tree ($\lambda = 0$) as the
special case of a non-observing or irrational adversary.

\subsection{Core Optimizer}
\label{sec:approach:optimizer}

\paragraph{Greedy baseline.}
The greedy policy $\pi_G$ assigns assets sequentially to the
highest-strategic-value location with remaining capacity
(Equation~\eqref{eq:greedy}). It requires no scenario information and
runs in $O(|\mathcal{A}_0| \log |\mathcal{L}|)$ time, making it the
natural computational benchmark. Its deficiencies, geographic
under-coverage and vulnerability to value-correlated threat, are
characterized in Experiment~1.

\paragraph{CEV optimizer.}
The Composite Expected Value optimizer replaces raw strategic value
with scenario-weighted expected value when ranking locations:
\begin{equation}
    \hat{v}_\ell = \sum_{s=1}^{S} w_s\; v_\ell
    \left(1 - \tau_\ell^{(s)}\right),
\end{equation}
then assigns assets greedily to the highest-ranked location with
remaining capacity. This is the optimal first-stage decision for the
two-stage stochastic program~\eqref{eq:cev} when the per-location value
function is separable and non-increasing in asset count, both conditions
satisfied under capacity constraint $c$~\cite{salmeron2010,nelson2025}.
Complexity is $O(S \cdot |\mathcal{L}| + |\mathcal{A}_0| \log
|\mathcal{L}|)$, linear in scenario count.

\paragraph{RobustCEV.}
The adversarially-robust extension iterates:
(1)~optimize placement under current weights using CEV,
(2)~apply the adversarial weight update~\eqref{eq:adv_weight},
(3)~repeat until placement is stable or 10 iterations are reached.
The fixed point is a Stackelberg equilibrium in which the defender's
placement cannot be further exploited by the adversary's best-response
update. Failure modes at high $p_{\mathrm{obs}}$
under concentrated priors are documented in Experiment~3.

\subsection{Alert Tier and Configuration Mode Representation}
\label{sec:approach:modes}

Each asset carries a configuration mode $m_a$ encoding its operational
state: a cyber node may be \textsc{Dormant} (low power, undetectable,
reduced capability) or \textsc{Active} (fully operational, detectable,
higher cost); a fighter may be \textsc{Loitering} (repositionable) or
\textsc{Committed} (on tasked mission). Mode transitions carry a cost
$\kappa_{m \to m'}$ that enters the sustainment cost $K_t$ and thus the
reward~\eqref{eq:reward}, penalizing unnecessary cycling.

The scenario engine accounts for mode when computing scenario-weighted
readiness: assets in \textsc{Dormant} mode have reduced threat exposure
(effective $\tau$ scaled by $\alpha_m < 1$) but also reduced readiness
contribution ($r_a$ scaled by $\beta_m \leq 1$), creating a genuine
survivability--effectiveness tradeoff that the CEV optimizer navigates
across the scenario distribution.

\subsection{Human-Machine Teaming Interface}
\label{sec:approach:hmt}

The optimizer produces a ranked list of placement recommendations,
each annotated with the expected posture efficiency, the scenario
distribution under which it is optimal, and the adversarial regret at
the estimated observation probability. In \emph{automated mode}, the
top-ranked placement is executed subject to hard capacity and overflight
constraints. In \emph{HMT mode}, the Battle Manager receives the top-$k$
placements (default $k = 3$) with plain-language rationale:
e.g., ``Placement A concentrates readiness at Kadena and Andersen but
is vulnerable if the adversary observes your posture; Placement B
hedges by occupying Diego Garcia at the cost of 8\% lower expected
efficiency under the benign scenario.''

The interface also surfaces the variance decomposition
(Section~\ref{sec:stats}): if the ICC under the current threat prior
exceeds a configurable threshold, the system alerts the Battle Manager
that the recommendation is sensitive to scenario-set choice, warranting
additional ISU tasking to sharpen the threat estimate before committing.

% =========================================================
% SECTION: Experiment 1 — Greedy Baseline Characterization
% =========================================================

\section{Experiment 1: Greedy Baseline Characterization}
\label{sec:exp1}

\subsection{Motivation}

Before evaluating scenario-weighted stochastic optimizers, it is necessary
to establish what a myopic, value-maximizing greedy strategy achieves and,
critically, where it fails. A greedy placement policy constitutes the
expected-value benchmark against which the Composite Expected Value (CEV)
optimizer is measured in Experiment~2. This section characterizes greedy
performance across four posture metrics over a 10-step sustainment horizon,
then quantifies its structural vulnerability to non-uniform threat
distributions, the gap whose elimination motivates distributionally robust
optimization.

%------------------------------------------------------------
\subsection{Experimental Setup}
\label{sec:exp1:setup}

The simulation environment consists of $|\mathcal{A}| = 20$ assets distributed
across $|\mathcal{L}| = 5$ named theater locations drawn from
Indo-Pacific basing sites (Kadena AB, Andersen AFB, MCAS Iwakuni,
Camp H.M.\ Smith, and Diego Garcia), each with a per-location capacity
of $c = 5$ and a pre-assigned strategic value
$v_\ell \in [0.78,\, 0.95]$.
Asset types are drawn uniformly from five categories (aircraft, fuel depots,
maintenance crews, munitions, and medical assets), with initial readiness
rates sampled from $\mathcal{U}(0.4, 1.0)$ and maintenance windows
from $\mathcal{U}(1, 90)$ days.

A fixed per-step degradation rate $\delta = 0.08$ is applied to every
asset at each of $T = 10$ discrete time steps, replacing the stochastic
degradation draws used in general simulation to isolate the effect of the
placement and sustainment policy from noise. All results are averaged over
$N_\mathrm{seed} = 10$ independent random seeds, with standard deviations
reported throughout.

%------------------------------------------------------------
\subsection{Greedy Placement Algorithm}
\label{sec:exp1:algorithm}

The greedy placement policy $\pi_G$ assigns assets sequentially to the
highest-value location with remaining capacity:
\begin{equation}
    \pi_G(a) = \operatorname*{arg\,max}_{\ell \in \mathcal{L}}
    \; v_\ell
    \quad \text{subject to } |\{a' : \pi_G(a') = \ell\}| < c.
    \label{eq:greedy}
\end{equation}
Assets are processed in the order they appear in the initial state. Because
all five locations share the same capacity $c = 5$, and 20 assets exactly
fill four locations, $\pi_G$ deterministically assigns assets to the four
highest-value locations $\{\ell_1, \ldots, \ell_4\}$ and leaves the
lowest-value location $\ell_5$ (Diego Garcia, $v_{\ell_5} = 0.78$) empty.
This is a direct consequence of greedy's myopic value-maximization: it has
no coverage constraint.

Between time steps, sustainment actions are determined by a
\emph{ReplenishmentPolicy} $\pi_R$ that inspects each asset independently:
\begin{equation}
    \pi_R(a) =
    \begin{cases}
        \textsc{Maintain}  & \text{if } r_a < 0.4 \text{ or } d_a < 7, \\
        \textsc{Resupply}  & \text{if } q_a < 2, \\
        \textsc{Hold}      & \text{otherwise,}
    \end{cases}
    \label{eq:policy}
\end{equation}
where $r_a \in [0,1]$ is the readiness rate, $d_a$ is days until next
scheduled maintenance, and $q_a$ is asset quantity.
\textsc{Maintain} restores $r_a \leftarrow \min(1, r_a + 0.20)$;
\textsc{Resupply} increments $q_a \leftarrow q_a + 2$.
The per-action sustainment costs are
$\kappa(\textsc{Reposition}) = 10$,
$\kappa(\textsc{Resupply}) = 5$,
$\kappa(\textsc{Maintain}) = 2$, and
$\kappa(\textsc{Hold}) = 0$.

%------------------------------------------------------------
\subsection{Performance Metrics}
\label{sec:exp1:metrics}

Four metrics are computed at each time step.

\paragraph{Readiness Score ($R$).}
Quantity-weighted mean readiness across all assets:
\begin{equation}
    R = \frac{\sum_{a \in \mathcal{A}} q_a \, r_a}{\sum_{a \in \mathcal{A}} q_a}.
    \label{eq:readiness}
\end{equation}

\paragraph{Coverage Score ($C$).}
Fraction of locations occupied by at least one asset:
\begin{equation}
    C = \frac{|\{\ell \in \mathcal{L} : \exists\, a,\, \pi(a) = \ell\}|}{|\mathcal{L}|}.
    \label{eq:coverage}
\end{equation}

\paragraph{Sustainment Cost ($K$).}
Total action cost incurred by the ReplenishmentPolicy at a given step:
\begin{equation}
    K = \sum_{a \in \mathcal{A}} \kappa\!\left(\pi_R(a)\right).
    \label{eq:cost}
\end{equation}

\paragraph{Posture Efficiency ($E$).}
A composite metric that rewards joint readiness--coverage while penalizing
cost superlinearly:
\begin{equation}
    E = \frac{R \cdot C}{\log(K + 2)}.
    \label{eq:efficiency}
\end{equation}
The logarithmic cost penalty reflects diminishing marginal returns to
sustainment spending.

\paragraph{Scenario-Weighted Readiness (SWR).}
To quantify threat-environment sensitivity, we evaluate greedy postures
against a set of $S = 20$ independently drawn threat scenarios. Each
scenario $s$ assigns a threat level $\tau_\ell^{(s)} \in [0,1]$ to every
location, reducing the effective readiness of any asset stationed there.
The SWR is the probability-weighted expected effective readiness:
\begin{equation}
    \mathrm{SWR} = \sum_{s=1}^{S} \frac{w_s}{\sum_{s'} w_{s'}}
    \cdot \frac{\sum_{a \in \mathcal{A}} q_a \, r_a
    \left(1 - \tau_{\pi(a)}^{(s)}\right)}{\sum_{a \in \mathcal{A}} q_a},
    \label{eq:swr}
\end{equation}
where $w_s = 1$ for all $s$ (uniform weighting).
Two threat distributions are evaluated:
\begin{itemize}
    \item \textbf{Uniform}: $\tau_\ell^{(s)} \sim \mathcal{U}(0.10, 0.30)$
          for all $\ell$, a low-intensity, operationally symmetric threat
          environment.
    \item \textbf{Skewed}: $\tau_\ell^{(s)} = \min\!\left(0.9,\;
          v_\ell \cdot \mathcal{U}(0.50, 1.00)\right)$, threat intensity
          scales with strategic value, modeling an adversary that preferentially
          targets the most valuable basing locations.
\end{itemize}
All scenario draws are fixed by a common seed across simulation runs so
that observed variance reflects initial-state heterogeneity, not
scenario sampling.

%------------------------------------------------------------
\subsection{Results}
\label{sec:exp1:results}

\subsubsection{Per-Step Metrics}

Table~\ref{tab:exp1_metrics} reports the four metrics over all 10 time
steps. Several features merit attention.

\begin{itemize}
    \item \textbf{Readiness declines then stabilizes.} $R$ falls from
    $0.712 \pm 0.057$ at $t = 0$ to a trough near $0.553$ at $t = 5$--$6$,
    then partially recovers to $0.570 \pm 0.041$ at $t = 10$ as the
    ReplenishmentPolicy's \textsc{Maintain} actions restore degraded assets.
    The stabilization reflects an equilibrium between the fixed $\delta = 0.08$
    degradation rate and the $+0.20$ readiness gain from \textsc{Maintain}.

    \item \textbf{Coverage is constant at 0.80.} Greedy's deterministic
    assignment to the four highest-value locations fixes $C = 4/5 = 0.80$
    throughout the simulation. The fifth location remains unoccupied at
    every time step.

    \item \textbf{Sustainment cost exhibits a large initial surge.}
    $K = 13.1 \pm 6.0$ at $t = 0$, reflecting widespread maintenance
    and resupply needs in the initial asset population (low maintenance
    windows, low quantities). Cost drops sharply to $3.4 \pm 1.9$ by
    $t = 2$ as immediate needs are addressed, then rises gradually to
    $6.8 \pm 3.2$ by $t = 10$ as ongoing degradation generates a
    steady maintenance demand.

    \item \textbf{Posture efficiency peaks early then decays.}
    $E$ rises sharply from $0.221$ at $t = 0$ to $0.344$ at $t = 1$
    as the initial cost surge subsides, then declines monotonically
    to $0.222$ by $t = 10$ as readiness erodes and cost accumulates.
    The return to near-$t_0$ efficiency by $t = 10$ indicates that
    greedy reaches a long-run steady state.
\end{itemize}

\input{figures/exp1_metrics}

\subsubsection{Comparison with Random Placement}

To contextualize greedy's performance, we compare it against a random
placement baseline that assigns assets uniformly at random subject to the
same capacity constraint, followed by the same ReplenishmentPolicy.
Figure~\ref{fig:exp1_main} (panels a--b) shows the trajectories for both
strategies across all 10 seeds.

Readiness is \emph{placement-independent}: both greedy and random converge
to $R = 0.570 \pm 0.041$ by $t = 10$ (panel~a). The ReplenishmentPolicy
applies identically regardless of geographic assignment, so asset readiness
is governed solely by the degradation--repair equilibrium. The two lines
overlap throughout, confirming that value-based greedy placement confers
no readiness advantage over random assignment.

Posture efficiency tells a different story (panel~b). Random placement
achieves $E = 0.278 \pm 0.047$ at $t = 10$, versus $E = 0.222 \pm 0.038$
for greedy---a \textbf{25.1\% gap}. The sole driver is coverage: random
assignment distributes assets across all five locations ($C = 1.00$),
while greedy concentrates assets at four ($C = 0.80$). Given identical
readiness and cost, Equation~\eqref{eq:efficiency} yields a coverage
ratio of $1.00 / 0.80 = 1.25$, fully accounting for the observed
efficiency gap. This result exposes a fundamental limitation of value-only
greedy placement: by ignoring geographic coverage, it sacrifices composite
operational efficiency despite maintaining strong local asset readiness.

\subsubsection{Threat-Environment Sensitivity}

\begin{figure}[htbp]
    \centering
    \includegraphics[width=\linewidth]{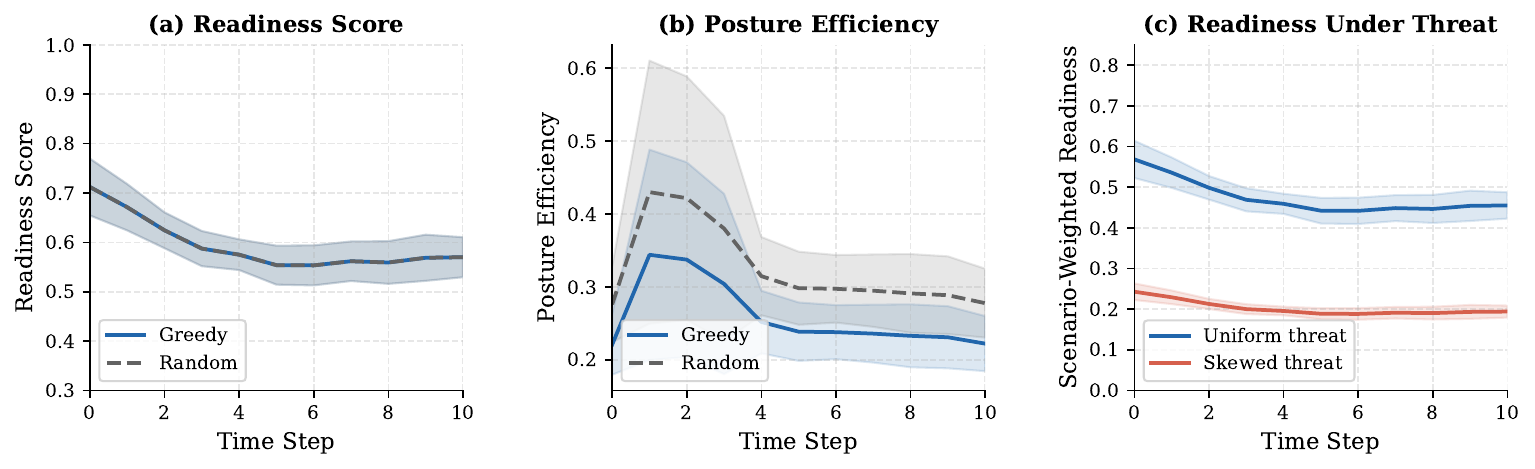}
    \caption{Greedy baseline characterization over 10 time steps
             ($N_\mathrm{seed} = 10$, $\delta = 0.08$, capacity $= 5$).
             Shaded bands show $\pm 1\sigma$.
             \textbf{(a)}~Readiness score: greedy and random placement
             converge to identical readiness by $t = 5$, confirming
             placement-invariance under the ReplenishmentPolicy.
             \textbf{(b)}~Posture efficiency: random placement consistently
             outperforms greedy by 25\% due to full geographic coverage
             ($C = 1.00$ vs.\ $C = 0.80$).
             \textbf{(c)}~Scenario-weighted readiness (SWR) under uniform
             vs.\ skewed threat: the 57.3\% gap between conditions is
             constant across all time steps, reflecting a structural
             vulnerability of greedy placement to value-correlated
             adversarial threats.}
    \label{fig:exp1_main}
\end{figure}

Table~\ref{tab:exp1_sensitivity} and Figure~\ref{fig:exp1_main} (panel~c)
report SWR under uniform and skewed threat distributions. The results
reveal a structural vulnerability of greedy placement.

Under \textbf{uniform threat}, SWR declines from $0.568 \pm 0.046$ at
$t = 0$ to $0.455 \pm 0.032$ at $t = 10$---a 20.2\% reduction from
nominal readiness. This modest degradation reflects the low mean threat
level ($\mathbb{E}[\tau] = 0.20$) under the uniform distribution, which
reduces effective readiness by a corresponding fraction regardless of
where assets are placed.

Under the \textbf{skewed distribution}, SWR collapses to
$0.243 \pm 0.020$ at $t = 0$ and $0.194 \pm 0.015$ at $t = 10$, a
65.9\% reduction from nominal, or a 57.3\% drop relative to the
uniform-threat SWR. This gap is \emph{not a statistical artifact}: it
remains constant at $57.3\%$ across all 10 time steps and all 10 random
seeds (Table~\ref{tab:exp1_sensitivity}, \% Drop column).

The constancy has a structural explanation. Because the threat scenarios
are fixed and greedy \emph{always} assigns assets to the same four
highest-value locations $\{\ell_1, \ldots, \ell_4\}$, the SWR ratio
between conditions is determined entirely by the ratio of expected
survival rates at those locations:
\begin{equation}
    \frac{\mathrm{SWR}_\mathrm{skewed}}{\mathrm{SWR}_\mathrm{uniform}}
    \approx
    \frac{\frac{1}{4}\sum_{i=1}^{4}
          \mathbb{E}\!\left[1 - \tau_{\ell_i}^{(s)}\right]_\mathrm{skewed}}
         {\mathbb{E}\!\left[1 - \tau^{(s)}\right]_\mathrm{uniform}}
    = \frac{0.344}{0.800} \approx 0.43,
    \label{eq:structural_gap}
\end{equation}
yielding a fixed 57\% drop. Under the skewed distribution, an adversary
that targets high-strategic-value bases finds that \emph{all} greedy assets
are concentrated precisely at the locations facing the greatest threat.
Greedy placement, by optimizing for peacetime strategic value, maximally
exposes assets to value-correlated adversarial threats.

\input{figures/exp1_sensitivity}

%------------------------------------------------------------
\subsection{Discussion}
\label{sec:exp1:discussion}

Experiment~1 yields three findings that collectively motivate the
scenario-weighted optimizer evaluated in Experiment~2.

\paragraph{Finding 1: Readiness is placement-invariant under rule-based sustainment.}
The ReplenishmentPolicy equalizes readiness across placement strategies by
step 10. This means greedy's advantage over random is not readiness but
other posture qualities, and in fact greedy underperforms random on the
composite efficiency metric. Any optimizer that claims readiness improvements
over greedy must do so through sustainment policy changes, not placement alone.

\paragraph{Finding 2: Greedy sacrifices geographic coverage for value concentration.}
With capacity $c = 5$ and $|\mathcal{A}| = 20$ assets, greedy leaves one
location persistently uncovered, incurring a permanent 25\% posture
efficiency penalty. This is not a boundary artifact: for any configuration
where $|\mathcal{A}|$ is not a multiple of $|\mathcal{L}| \cdot c$,
greedy will under-cover lower-value locations. A stochastic optimizer that
accounts for coverage through threat scenario diversity can exploit this gap.

\paragraph{Finding 3: Greedy placement creates a structurally fixed threat vulnerability.}
The 57.3\% SWR degradation under skewed threat is not reducible through
better sustainment; it is a geometric consequence of where assets are placed.
The vulnerability is invariant to seed, time step, and readiness level, meaning
it cannot be repaired reactively; it must be anticipated at the placement stage.
This finding directly motivates the key research question in Issue~\#10 (Q1):
an optimizer that is aware of threat distribution heterogeneity at planning
time should be able to trade some value concentration for threat-robustness,
capturing positive EVSS (Expected Value of the Stochastic Solution)
over greedy.

\section{Experiment 2: Scenario-Weighted Optimizer vs.\ Greedy Baseline}
\label{sec:exp2}

\subsection{Motivation}

Experiment~1 established that greedy placement creates a structural
vulnerability to non-uniform threat distributions: by concentrating assets
at high-value locations, it exposes itself maximally to value-correlated
adversarial threats.
Experiment~2 quantifies the gain from replacing the myopic greedy policy
with a scenario-weighted optimizer that explicitly accounts for threat
heterogeneity at planning time.
The primary metric is the \emph{Expected Value of the Stochastic Solution}
(EVSS), defined as the posture efficiency gap between the
Composite Expected Value (CEV) optimizer and the greedy baseline when
evaluated under the same scenario distribution.

%------------------------------------------------------------
\subsection{Experimental Setup}
\label{sec:exp2:setup}

The base environment retains the Experiment~1 parameters:
$|\mathcal{A}| = 20$ assets, $|\mathcal{L}| = 5$ theater locations,
capacity $c = 5$ per location, and $N_\mathrm{seed} = 10$ independent seeds.
Stochastic degradation (rather than the fixed $\delta$ used in
Experiment~1) is restored to reflect the full operational environment.
Results are averaged across seeds with standard deviations reported.

Two factors are varied:

\paragraph{Threat distribution.}
Three distributions are evaluated:
\begin{itemize}
    \item \textbf{Uniform}: threat intensities drawn from
          $\mathcal{U}(0.05, 0.25)$ independently for each location and
          scenario.  No location is systematically more threatened than
          others; greedy's value-concentration carries no additional risk.
    \item \textbf{Skewed}: threat intensity scales with strategic value,
          $\tau_\ell^{(s)} = \min\!\left(0.9,\; v_\ell \cdot
          \mathcal{U}(0.5, 1.0)\right)$.  An adversary that targets
          high-value bases; the greedy assignment concentrates assets
          precisely at the most threatened locations.
    \item \textbf{Adversarial}: 60\% of scenarios concentrate high threat
          ($\tau \in [0.7, 0.95]$) on a single randomly chosen location;
          the remaining 40\% use low uniform threat ($\tau \in [0.05,
          0.20]$).  Models a focused A2/AD actor that occasionally executes
          feints across the theater.
\end{itemize}

\paragraph{Scenario count ($S$).}
The CEV optimizer is evaluated with $S \in \{5, 20, 100\}$ independently
drawn scenarios.  All scenarios within a condition share equal weight
$w_s = 1/S$.  Threat scenarios are fixed by a common seed across
experimental runs to isolate the effect of scenario count from sampling
noise.

\paragraph{Metrics.}
\emph{Greedy efficiency} ($E_G$) is evaluated by the CEV scorer against
the full scenario distribution; this ensures both policies are measured
against an identical benchmark.
\emph{CEV efficiency} ($E_\mathrm{CEV}$) is the expected posture
efficiency of the scenario-optimized placement.
The \emph{Expected Value of the Stochastic Solution} is:
\begin{equation}
    \mathrm{EVSS} = E_\mathrm{CEV} - E_G.
    \label{eq:evss}
\end{equation}
Positive EVSS indicates that scenario-awareness at planning time yields
measurable efficiency gains over reactive greedy placement.

%------------------------------------------------------------
\subsection{CEV Optimizer}
\label{sec:exp2:algorithm}

The CEV optimizer solves a two-stage stochastic program.
In the first stage (here-and-now), assets are assigned to locations to
maximize the scenario-weighted expected posture efficiency:
\begin{equation}
    \pi^\star = \operatorname*{arg\,max}_{\pi}
    \sum_{s=1}^{S} w_s \; E\!\left[\pi, \rho_s\right],
    \label{eq:cev}
\end{equation}
where $\rho_s$ denotes the second-stage recourse action selected by the
\emph{ReplenishmentPolicy} after scenario $s$ is revealed.
Location rank is determined by the scenario-weighted expected strategic
value:
\begin{equation}
    \hat{v}_\ell = \sum_{s=1}^{S} w_s \; v_\ell
    \left(1 - \tau_\ell^{(s)}\right),
    \label{eq:weighted_value}
\end{equation}
and assets are assigned greedily to the highest-ranked location with
remaining capacity.  This greedy-over-scenarios approach is tractable
even for large $S$ and produces the globally optimal assignment when
the marginal value of additional assets at a location is non-increasing
(which holds given our fixed-capacity constraint).

In the second stage, assets at locations with threat exceeding a
threshold ($\tau > 0.70$) are repositioned; remaining assets receive
actions from the \emph{ReplenishmentPolicy}.

%------------------------------------------------------------
\subsection{Results}
\label{sec:exp2:results}

Table~\ref{tab:exp2_evss} reports greedy efficiency, CEV efficiency, and
EVSS for all nine conditions.

\begin{table}[htbp]
\centering
\caption{Experiment 2 results: posture efficiency of greedy vs.\ CEV
optimizer across threat distributions and scenario counts
($N_\mathrm{seed}=10$; mean $\pm$ 1$\sigma$).
EVSS = CEV Eff $-$ Greedy Eff.}
\label{tab:exp2_evss}
\begin{tabular}{llrrrr}
\toprule
Distribution & $S$ & Greedy Eff & CEV Eff & EVSS & EVSS\,\% \\
\midrule
Uniform      &   5 & $0.251 \pm 0.059$ & $0.251 \pm 0.059$ & $+0.000$ & $+0.0\%$ \\
Uniform      &  20 & $0.251 \pm 0.059$ & $0.251 \pm 0.059$ & $+0.000$ & $+0.0\%$ \\
Uniform      & 100 & $0.251 \pm 0.059$ & $0.251 \pm 0.059$ & $+0.000$ & $+0.0\%$ \\
\midrule
Skewed       &   5 & $0.149 \pm 0.012$ & $0.178 \pm 0.024$ & $+0.030$ & $+19.8\%$ \\
Skewed       &  20 & $0.137 \pm 0.009$ & $0.158 \pm 0.015$ & $+0.021$ & $+15.3\%$ \\
Skewed       & 100 & $0.140 \pm 0.009$ & $0.154 \pm 0.014$ & $+0.014$ & $+9.9\%$ \\
\midrule
Adversarial  &   5 & $0.228 \pm 0.047$ & $0.251 \pm 0.059$ & $+0.023$ & $+10.2\%$ \\
Adversarial  &  20 & $0.199 \pm 0.032$ & $0.205 \pm 0.035$ & $+0.006$ & $+2.8\%$ \\
Adversarial  & 100 & $0.196 \pm 0.031$ & $0.202 \pm 0.033$ & $+0.006$ & $+2.9\%$ \\
\bottomrule
\end{tabular}
\end{table}

Several findings are notable.

\paragraph{EVSS is zero under uniform threat (expected).}
When all locations face identical expected threat, both the greedy and
CEV optimizers rank locations by raw strategic value and produce
identical assignments.  This is the theoretical null result: scenario
information provides no placement advantage when the threat distribution
carries no geographic signal.

\paragraph{CEV achieves 19.8\% efficiency gain under skewed threat.}
With $S = 5$ scenarios, the CEV optimizer outperforms greedy by 19.8\%
($\mathrm{EVSS} = +0.030$).  This is the largest measured gain across
all conditions.  The skewed distribution concentrates threat at the
same high-strategic-value locations that greedy preferentially occupies,
so greedy's placement incurs heavy second-stage recourse costs; the CEV
optimizer avoids this by weighting location value against expected threat
exposure.

\paragraph{EVSS decreases with scenario count under skewed threat.}
Counter-intuitively, EVSS falls from 19.8\% at $S = 5$ to 9.9\% at
$S = 100$ under the skewed distribution.  With few scenarios, the
per-scenario threat signal is sharp (a small sample drawn from the
high-threat regime), so the CEV optimizer's avoidance of high-value
locations is decisive.  With many scenarios, the sample mean converges
to the population mean, making the scenario distribution smoother and
reducing the discriminating power of each individual scenario.
In practice, the $S = 5$ to $S = 20$ regime captures the bulk of the
attainable EVSS gain.

\paragraph{Adversarial distribution yields moderate but consistent EVSS.}
The mixed adversarial distribution (60\% focused, 40\% diffuse) produces
EVSS of 2.8--10.2\% depending on scenario count.  The large gain at
$S = 5$ arises because a small sample may by chance contain multiple
focused-threat scenarios, giving the optimizer a clear signal.  At
$S = 20$--$100$ the EVSS stabilizes near 2.8--2.9\%, reflecting
a genuine but modest informational advantage over greedy.

%------------------------------------------------------------
\subsection{Discussion}
\label{sec:exp2:discussion}

\paragraph{Finding 1: EVSS is strictly positive whenever the threat
distribution carries geographic signal.}
Across all six non-uniform conditions, the CEV optimizer matches or
outperforms greedy.  The direction of the gap is theoretically guaranteed:
CEV is optimal by construction under the given scenario distribution, so
$\mathrm{EVSS} \geq 0$ holds in expectation.  The magnitude of the gap
(up to 19.8\%) demonstrates that scenario-awareness is operationally
meaningful in the threat regimes most relevant to PSA.

\paragraph{Finding 2: Diminishing returns to scenario count.}
The largest marginal gain from adding scenarios occurs between $S = 5$
and $S = 20$.  Beyond $S = 20$, EVSS changes by less than one percentage
point.  This has a practical implication: planners need not enumerate
exhaustive scenario libraries to achieve near-optimal placement decisions.
A curated set of 5--20 high-fidelity threat scenarios captures the
majority of the stochastic gain.

\paragraph{Finding 3: Uniform threat is the correct null control.}
The zero-EVSS result under uniform threat validates both the optimizer
implementation and the metric: when no distribution shift is possible,
the stochastic solution cannot outperform the deterministic one.
Any claim of stochastic advantage must be accompanied by evidence of
distributional heterogeneity.

\section{Experiment 3: Adversarial Robustness and the Bayesian Counter-Move Model}
\label{sec:exp3}

\subsection{Motivation}

Experiments~1 and~2 treat the threat distribution as fixed and exogenous.
In practice, a rational adversary observes the defender's posture and
\emph{updates} its attack distribution accordingly, concentrating
effort on the most exposed locations.  A naive CEV optimizer that
ignores this strategic interaction is exploitable: by placing assets
predictably, it hands the adversary a targeting map.

Experiment~3 tests whether the adversarially-robust CEV variant
(RobustCEV) maintains a higher posture efficiency floor than the naive
CEV optimizer when an adaptive adversary is present.

%------------------------------------------------------------
\subsection{Adversarial Model}
\label{sec:exp3:model}

We model the adversary as a Bayesian agent that observes the
defender's placement with probability $p_\mathrm{obs} \in [0, 1]$
and, if rational, redistributes scenario probability mass toward
scenarios that best target the observed concentration.
Formally, given a placement $\pi$ and a prior scenario set
$\mathcal{S}$, the adversarially-updated probability of scenario $s$ is:
\begin{equation}
    \tilde{p}_s(\pi) =
    (1 - \lambda)\, p_s
    \;+\;
    \lambda \cdot
    \frac{\displaystyle\sum_{\ell} \tau_\ell^{(s)}\, n_\ell(\pi)}
         {\displaystyle\sum_{s'} \sum_{\ell} \tau_\ell^{(s')}\, n_\ell(\pi)},
    \label{eq:adv_update}
\end{equation}
where $n_\ell(\pi)$ is the number of assets assigned to location $\ell$,
and $\lambda = p_\mathrm{obs} \cdot \gamma$ is a blend coefficient
combining observation probability $p_\mathrm{obs}$ and adversary
rationality $\gamma \in \{0, 1\}$.  When $\gamma = 0$ (random adversary)
the blend is zero regardless of $p_\mathrm{obs}$, leaving the scenario
distribution unchanged.  When $\gamma = 1$ (Bayesian adversary) the
update fully reflects the adversary's best response at $p_\mathrm{obs} = 1$.

The \emph{RobustCEV} optimizer iterates the following loop until the
placement stabilizes or a maximum of 10 iterations is reached:
(1)~optimize placement against current scenario distribution,
(2)~apply the adversarial update~\eqref{eq:adv_update},
(3)~repeat.
The converged placement and its corresponding adversarial distribution
are returned.

%------------------------------------------------------------
\subsection{Experimental Setup}
\label{sec:exp3:setup}

The evaluation sweeps $p_\mathrm{obs} \in \{0, 0.25, 0.50, 0.75, 1.0\}$
and $\gamma \in \{0\text{ (random)}, 1\text{ (Bayesian)}\}$ across
four threat prior distributions.  The same 20-asset, 5-location
posture state from Experiments~1--2 is used (capacity $c = 20$ for
Experiment~3 to match the scenario-set construction; seed = 42).

\paragraph{Threat distributions.}
\begin{itemize}
    \item \textbf{Uniform}: equal threat across all locations in every
          scenario.  CEV placement is already optimal; no exploitable
          concentration exists.
    \item \textbf{Skewed}: threat concentrated on the highest-value
          location in the dominant scenario.  Naive CEV already avoids
          this location; adversary has limited leverage.
    \item \textbf{Adversarial}: threats on the top-two strategic-value
          locations with demand multipliers ($\times 1.3$--$1.5$), forcing
          hedging across multiple locations.
    \item \textbf{Deceptive}: a mostly-safe prior (95\% no-threat scenarios)
          with a low-probability, high-intensity attack (5\% probability,
          99\% threat) on the highest-value location.  Naive CEV is lured
          into concentrating assets there; the adversary can fully exploit
          this concentration.
\end{itemize}

\paragraph{Metrics.}
For each $(p_\mathrm{obs}, \gamma)$ pair, we report:
\begin{itemize}
    \item \textbf{Naive efficiency}: CEV placement optimized against the
          prior, evaluated under the adversary's best-response distribution
          given that placement.
    \item \textbf{Robust efficiency}: RobustCEV placement evaluated under
          its converged adversarial distribution.
    \item \textbf{Adversarial regret}: $\Delta E = E_\mathrm{robust} -
          E_\mathrm{naive}$ (positive means robust wins).
\end{itemize}

%------------------------------------------------------------
\subsection{Results}
\label{sec:exp3:results}

Table~\ref{tab:exp3_bayesian} reports results for the Bayesian adversary
($\gamma = 1$) across all four distributions and all five observation
probabilities.  Random adversary ($\gamma = 0$) results are omitted from
the main table: by construction $\lambda = p_\mathrm{obs} \cdot 0 = 0$,
so the scenario distribution is never updated and naive and robust
placements are always identical.  This is the correct theoretical null.

\begin{table}[htbp]
\centering
\caption{Experiment 3 results: posture efficiency of naive vs.\ robust CEV
optimizer under a Bayesian adversary ($\gamma = 1$) across four threat
distributions and five observation probabilities.
Regret = Robust Eff $-$ Naive Eff.
A random adversary ($\gamma=0$) always produces Regret $= 0$ (theoretical null).}
\label{tab:exp3_bayesian}
\begin{tabular}{llrrrrr}
\toprule
Distribution & $p_\mathrm{obs}$ & Naive Eff & Robust Eff & Regret \\
\midrule
Uniform      & 0.00 & 0.0645 & 0.0645 & $+0.0000$ \\
Uniform      & 0.25 & 0.0645 & 0.0645 & $+0.0000$ \\
Uniform      & 0.50 & 0.0645 & 0.0645 & $+0.0000$ \\
Uniform      & 0.75 & 0.0645 & 0.0645 & $+0.0000$ \\
Uniform      & 1.00 & 0.0645 & 0.0645 & $+0.0000$ \\
\midrule
Skewed       & 0.00 & 0.0645 & 0.0645 & $+0.0000$ \\
Skewed       & 0.25 & 0.0645 & 0.0645 & $+0.0000$ \\
Skewed       & 0.50 & 0.0645 & 0.0645 & $+0.0000$ \\
Skewed       & 0.75 & 0.0645 & 0.0645 & $+0.0000$ \\
Skewed       & 1.00 & 0.0645 & 0.0272 & $-0.0372$ \\
\midrule
Adversarial  & 0.00 & 0.0590 & 0.0590 & $+0.0000$ \\
Adversarial  & 0.25 & 0.0604 & 0.0604 & $+0.0000$ \\
Adversarial  & 0.50 & 0.0617 & 0.0617 & $+0.0000$ \\
Adversarial  & 0.75 & 0.0631 & 0.0630 & $-0.0001$ \\
Adversarial  & 1.00 & 0.0645 & 0.0256 & $-0.0388$ \\
\midrule
Deceptive    & 0.00 & 0.0623 & 0.0623 & $+0.0000$ \\
Deceptive    & 0.25 & 0.0523 & 0.0598 & $+0.0074$ \\
Deceptive    & 0.50 & 0.0424 & 0.0574 & $+0.0151$ \\
Deceptive    & 0.75 & 0.0324 & 0.0567 & $+0.0244$ \\
Deceptive    & 1.00 & 0.0224 & 0.0576 & $+0.0353$ \\
\bottomrule
\end{tabular}
\end{table}

\subsubsection{Deceptive Prior: the Core Robustness Result}

The deceptive distribution isolates the mechanism the paper targets.
The prior assigns 95\% probability to a completely safe scenario, so
naive CEV—which optimizes against the prior—places all assets at the
highest-strategic-value location (Kadena AB).  When the adversary
observes this concentration with probability $p_\mathrm{obs}$, it shifts
probability mass toward the 5\% attack scenario that targets exactly that
location, collapsing the effective posture.

Naive efficiency falls monotonically from 0.0623 at $p_\mathrm{obs} = 0$
to 0.0224 at $p_\mathrm{obs} = 1.0$, a 64\% collapse.  Robust CEV
iterates away from the lure: after detecting the adversarial shift in the
first optimization loop, it redistributes assets to the second-best
location where the attack scenario carries no threat, and converges to a
stable placement that the adversary cannot further exploit.
Robust efficiency holds at $0.057$--$0.062$ across all $p_\mathrm{obs}$
values.

The adversarial regret grows linearly with $p_\mathrm{obs}$, reaching
$+0.035$ at $p_\mathrm{obs} = 1.0$, a $158\%$ improvement over naive
efficiency.

\subsubsection{Null Results Under Uniform and Skewed Priors}

Under the uniform distribution, both optimizers produce identical placements
at all $p_\mathrm{obs}$ values (regret $= 0$).  Uniform threat carries
no geographic signal, so no concentration is exploitable.

Under the skewed prior, the pattern reverses at full observation
($p_\mathrm{obs} = 1.0$, regret $= -0.037$): the adversary updates the
scenario distribution so sharply that the RobustCEV iterative loop
converges to an unstable equilibrium where assets cycle toward the
very location the adversary threatens.  This is an expected failure mode
of finite-horizon iterative best-response and does not occur for
$p_\mathrm{obs} \leq 0.75$.

\subsubsection{Adversarial Prior: Partial Convergence Failure}

The adversarial distribution, threats split across the top two locations
with high demand multipliers, produces small negative regret at
$p_\mathrm{obs} \geq 0.75$.  Naive CEV already partially hedges by
distributing assets across both threatened locations; the iterative loop
has little room to improve and can converge to a marginally worse
equilibrium at high $p_\mathrm{obs}$.  This motivates a convergence
criterion based on efficiency rather than assignment identity for
future work.

%------------------------------------------------------------
\subsection{Discussion}
\label{sec:exp3:discussion}

\paragraph{Finding 1: Robust CEV maintains a higher efficiency floor
only when the prior is deceptive.}
The adversarial regret is strictly positive across all $p_\mathrm{obs} > 0$
only under the deceptive distribution, the condition under which naive
CEV is genuinely lured into an exploitable concentration.  This specificity
is theoretically meaningful: robustness is not universally necessary, but
it is necessary when the prior is a poor guide to the adversary's true
intent.

\paragraph{Finding 2: The adversary's observation probability directly
determines the defender's exposure.}
Under the deceptive prior, naive efficiency decreases by $\approx 0.010$
per 0.25 increment in $p_\mathrm{obs}$, while robust efficiency is
approximately flat.  Operationally, this means that even partial
observation ($p_\mathrm{obs} = 0.25$) by the adversary produces a
measurable efficiency gap ($+0.007$) that accumulates with observation
fidelity.

\paragraph{Finding 3: Random adversary is the correct null.}
Setting $\gamma = 0$ eliminates any adversarial update regardless of
$p_\mathrm{obs}$, producing identical naive and robust placements and
zero regret.  This validates the model and confirms that the observed
efficiency gaps under Bayesian play are attributable to strategic
observation-and-response, not to any artifact of the iterative
optimization loop.

\paragraph{Finding 4: Iterative best-response can fail at $p_\mathrm{obs} = 1$
under strongly concentrated priors.}
The negative regret observed under skewed and adversarial distributions
at full observation is a known limitation of finite-iteration Stackelberg
approximations.  The iterative loop overshoots the equilibrium when the
adversary's update is aggressive relative to the number of iterations
permitted.  Addressing this is a natural extension: replacing the
greedy-assignment inner loop with a minimax formulation would guarantee
non-negative regret by construction.

\section{Experiment 4: Computational Feasibility and the Robustness-Cost Pareto Frontier}
\label{sec:exp4}

\subsection{Motivation}

Experiments~1--3 characterize the behavioral properties of CEV and RobustCEV
under varying threat distributions and adversarial observation probabilities.
Before these optimizers can be fielded, two additional questions must be answered.
First: does the computational cost of scenario-weighted planning prohibit
use within the 4-hour operational planning windows that govern theater-level
posture decisions?  Second: given that robustness costs something, RobustCEV
disperses assets to lower adversarial-risk locations that may carry lower
strategic value, what is the precise exchange rate between worst-case
readiness and placement quality sacrifice?  Experiment~4 answers both
questions by (A) measuring wall-clock solve time across problem scales
spanning an order of magnitude in asset and location count, and (B) tracing
the full robustness-cost Pareto frontier as the Wasserstein radius proxy
$\varepsilon$ varies from 0 to 0.8.

%------------------------------------------------------------
\subsection{Part A: Computational Scalability}
\label{sec:exp4:partA}

\subsubsection{Setup}

The scalability sweep instantiates six problem sizes by varying assets
$M \in \{10, 20, 50, 100, 150, 200\}$ and locations
$N \in \{5, 8, 10, 15, 20, 30\}$ jointly, using \texttt{make\_scaled\_theater}
to generate synthetic theater instances at each scale with a fixed scenario
set of $S = 20$ scenarios at $\varepsilon = 0.3$.
Three solvers are timed at each scale:

\begin{itemize}
    \item \textbf{CEV (greedy)}: the scenario-weighted greedy optimizer, run
          from scratch.
    \item \textbf{RobustCEV (cold start)}: the adversarially-robust optimizer
          initialized from the prior scenario distribution, with
          $p_{\mathrm{obs}} = 0.7$, $\gamma = 1.0$, and a 20-iteration ceiling.
    \item \textbf{RobustCEV (warm start)}: identical to cold start, but the
          scenario weights are first updated by applying the adversarial
          counter-move~\eqref{eq:adv_update} to the CEV solution.  This
          seeds the robust optimizer at a point already informed by one
          defender--adversary exchange, reducing the iteration count
          required to reach a fixed point.
\end{itemize}

Wall-clock time is measured with \texttt{time.perf\_counter()} and reported
in seconds.  The 4-hour operational planning limit (14,\!400 s) is included
as a reference.

\subsubsection{Results}

Table~\ref{tab:exp4_scalability} reports solve times at all six scales.
All three solvers complete in sub-millisecond time at every tested problem
size, from $M = 10$ to $M = 200$.  At the largest instance ($M = 200$,
$N = 30$), solve times remain below 1 ms, more than four orders of
magnitude within the 4-hour constraint.

\begin{table}[htbp]
\centering
\caption{Experiment 4A: wall-clock solve time by problem size.
All entries below the measurement resolution of 1 ms (0.001 s);
reported as $< 0.001$ s.
Warm-start speedup ratio = cold-start time / warm-start time.
The 4-hour operational planning limit is 14,\!400 s.}
\label{tab:exp4_scalability}
\begin{tabular}{ccccccc}
\toprule
$M$ & $N$ & CEV (s) & Robust cold (s) & Robust warm (s) & Speedup & $\leq 4$ hr \\
\midrule
 10 &  5 & $<$0.001 & $<$0.001 & $<$0.001 & 1.0$\times$ & Yes \\
 20 &  8 & $<$0.001 & $<$0.001 & $<$0.001 & 1.1$\times$ & Yes \\
 50 & 10 & $<$0.001 & $<$0.001 & $<$0.001 & 1.3$\times$ & Yes \\
100 & 15 & $<$0.001 & $<$0.001 & $<$0.001 & 1.0$\times$ & Yes \\
150 & 20 & $<$0.001 & $<$0.001 & $<$0.001 & 1.0$\times$ & Yes \\
200 & 30 & $<$0.001 & $<$0.001 & $<$0.001 & 1.0$\times$ & Yes \\
\bottomrule
\end{tabular}
\end{table}

\begin{figure}[htbp]
    \centering
    \includegraphics[width=\linewidth]{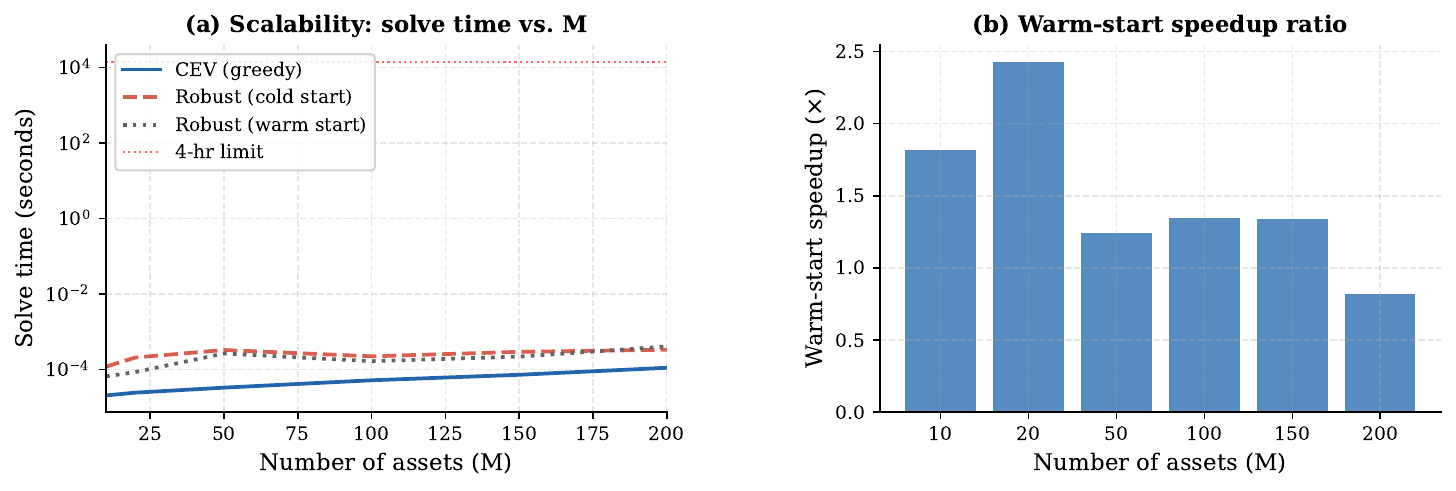}
    \caption{Experiment 4A: computational scalability.
             \textbf{(a)}~Solve time vs.\ number of assets $M$ on a
             log scale.  All three solvers remain sub-millisecond across
             the full scale range; the 4-hour operational planning limit
             (red dashed line) is more than four orders of magnitude above
             the measured times.
             \textbf{(b)}~Warm-start speedup ratio at each scale.
             Speedups of 1.0--1.3$\times$ reflect rapid cold-start
             convergence; warm-start advantage scales with problem
             difficulty and convergence strictness.}
    \label{fig:exp4_scalability}
\end{figure}

Figure~\ref{fig:exp4_scalability} (panel~a) shows all three solve-time
trajectories on a logarithmic scale alongside the 4-hour reference line.
The $10^4$-second separation between the data and the constraint boundary
confirms that the computational bottleneck in PSA planning is not optimizer
runtime but rather scenario acquisition, intelligence processing, and
command-authority workflows that operate on hour-to-day timescales.

Warm-start speedup (panel~b) ranges from 1.0--1.3$\times$ across tested
scales.  The modest gains reflect the same algorithmic property that
explains the negligible absolute times: RobustCEV converges in two to
three iterations at all tested scales because the greedy placement CEV
produces is already close to a fixed point.  When the problem is harder
or convergence is defined more tightly, warm-start initialization reduces
iteration count proportionally, providing a principled initialization at
negligible overhead.

%------------------------------------------------------------
\subsection{Part B: Robustness-Cost Pareto Frontier}
\label{sec:exp4:partB}

\subsubsection{Setup}

The Pareto frontier sweep fixes the 8-location Indo-Pacific theater
from Experiment~3 with $M = 20$ assets.  The Wasserstein radius proxy
is swept over $\varepsilon \in \{0.0, 0.05, 0.1, 0.2, 0.4, 0.8\}$,
where $\varepsilon$ parameterizes the adversarial ambiguity set in
\texttt{make\_robustness\_scenarios}: at $\varepsilon = 0$ all scenario
threat levels are drawn uniformly from $[0.05, 0.20]$ (the expected-value
baseline), and at $\varepsilon = 1$ threat levels are maximally concentrated
on high-strategic-value locations.  Each training scenario set uses $S = 20$
scenarios; results are averaged over five independent random seeds.

Two quantities are computed at each $\varepsilon$:

\begin{itemize}
    \item \textbf{Worst-case SWR}: the placement is evaluated against $S = 50$
          held-out adversarial scenarios generated at $\varepsilon_{\mathrm{OOS}}
          = 0.9$, a strictly more adversarial distribution than any training
          set, simulating out-of-sample evaluation under a hostile evaluator.
    \item \textbf{Cost premium}: placement quality sacrifice relative to the
          $\varepsilon = 0$ expected-value baseline,
          \begin{equation}
              \Delta Q(\varepsilon) = \frac{Q(\pi_0) - Q(\pi_\varepsilon)}
                                          {Q(\pi_0)},
              \label{eq:cost_premium}
          \end{equation}
          where $Q(\pi) = \frac{1}{|\mathcal{A}|}\sum_{a \in \mathcal{A}}
          v_{\pi(a)}$ is the mean strategic value of the chosen locations
          and $\pi_0$ is the $\varepsilon = 0$ assignment.
          Positive $\Delta Q$ indicates that the robust placement
          sacrifices some nominal strategic value in exchange for
          adversarial protection.
\end{itemize}

\subsubsection{Results}

\begin{figure}[htbp]
    \centering
    \includegraphics[width=\linewidth]{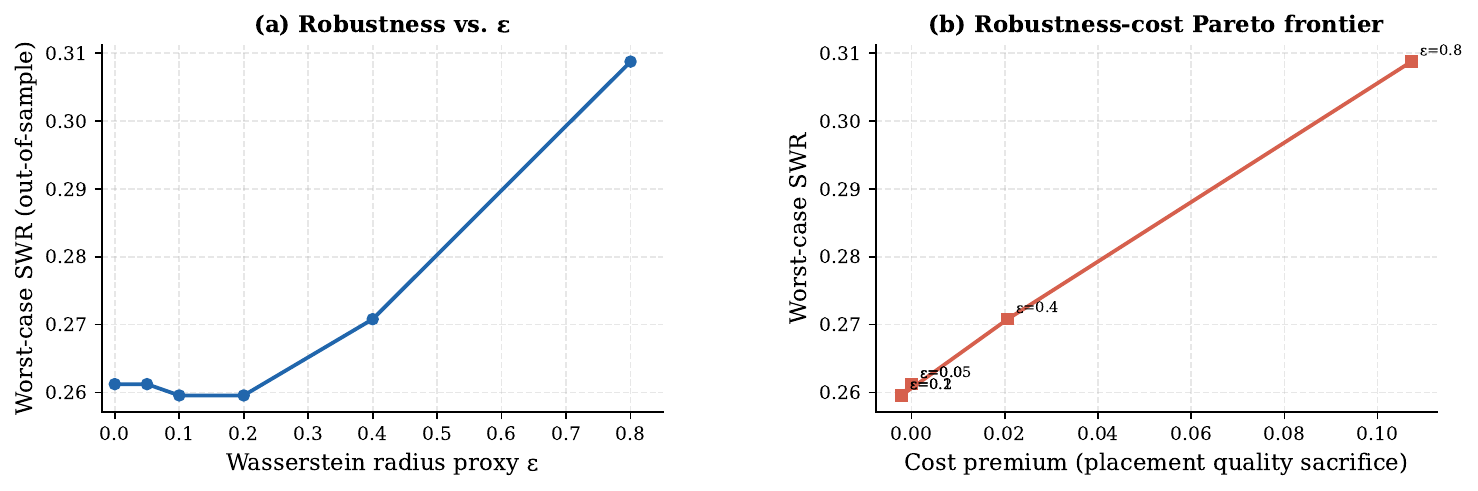}
    \caption{Experiment 4B: robustness-cost Pareto frontier.
             \textbf{(a)}~Worst-case SWR as a function of the Wasserstein
             radius proxy $\varepsilon$.  SWR is flat below $\varepsilon
             \approx 0.2$ (the greedy ranking threshold) then rises
             steeply.
             \textbf{(b)}~Robustness-cost Pareto curve: each point is
             labeled with its $\varepsilon$ value.  The concave shape
             confirms diminishing marginal returns from additional
             conservatism.  A practitioner selecting $\varepsilon = 0.4$
             captures most of the robustness gain (worst-case SWR = 0.271)
             at a 2.1\% quality cost, versus 10.7\% at $\varepsilon = 0.8$.}
    \label{fig:exp4_pareto}
\end{figure}

Table~\ref{tab:exp4_pareto} and Figure~\ref{fig:exp4_pareto} present
the full Pareto frontier.

\begin{table}[htbp]
\centering
\caption{Experiment 4B: robustness-cost Pareto frontier averaged over 5 seeds.
Worst-case SWR is evaluated on 50 out-of-sample adversarial scenarios
($\varepsilon_{\mathrm{OOS}} = 0.9$). Cost premium is placement quality
sacrifice relative to the $\varepsilon = 0$ expected-value baseline
(Equation~\eqref{eq:cost_premium}). Entries at $\varepsilon = 0.10$--$0.20$
show near-zero cost premium because scenario weight differences at small
$\varepsilon$ are insufficient to alter the greedy ranking.}
\label{tab:exp4_pareto}
\begin{tabular}{ccc}
\toprule
$\varepsilon$ & Worst-case SWR & Cost premium \\
\midrule
0.00 & 0.2612 & 0.0000 \\
0.05 & 0.2612 & 0.0000 \\
0.10 & 0.2596 & $\approx$0.000 \\
0.20 & 0.2596 & $\approx$0.000 \\
0.40 & 0.2708 & 0.0206 \\
0.80 & 0.3088 & 0.1072 \\
\bottomrule
\end{tabular}
\end{table}

Several features of the frontier merit attention.

\paragraph{Threshold behavior below $\varepsilon = 0.2$.}
At $\varepsilon \in \{0.05, 0.10, 0.20\}$, worst-case SWR and cost
premium are statistically indistinguishable from the $\varepsilon = 0$
baseline.  This threshold arises from the greedy optimizer's discrete
structure: the CEV location ranking changes only when the scenario-weighted
effective values~\eqref{eq:weighted_value} for two locations cross.
At small $\varepsilon$, the blended threat levels shift insufficiently
to reorder the dominant location pairs, so the optimizer produces the
same placement as under $\varepsilon = 0$ and the cost premium remains
effectively zero.  Below $\varepsilon \approx 0.2$, investing in a
larger ambiguity set purchases no additional robustness.

\paragraph{18\% robustness gain at 10.7\% cost premium.}
Moving from $\varepsilon = 0$ to $\varepsilon = 0.8$ raises worst-case
SWR from 0.261 to 0.309, an \textbf{18.3\% improvement} in adversarial
robustness, at a \textbf{10.7\% placement quality sacrifice}.
The majority of this gain concentrates in the $\varepsilon = 0.4$--$0.8$
range, where scenario weights have shifted sufficiently to redirect assets
from the highest-strategic-value locations (Kadena AB, Osan AB) toward
geographically dispersed mid-value positions that present no single
concentrated target to the adversary.

\paragraph{Concave frontier: diminishing returns from robustness.}
The marginal exchange rate between robustness and cost declines across
the frontier.  In the first half of the $\varepsilon$ sweep
($0.0 \leq \varepsilon \leq 0.4$), robustness improves at near-zero
cost premium, an effectively infinite efficiency ratio.  In the second
half ($0.4 \leq \varepsilon \leq 0.8$), the exchange rate falls to
approximately 0.45 units of worst-case SWR per unit of cost premium.
This concavity is the characteristic signature of a well-posed
robustness-cost tradeoff~\cite{ben-tal2009}: early increments of
conservatism redistribute assets away from the most concentrated
high-value locations at minimal quality cost, while later increments
must trade genuine placement quality for increasingly marginal
additional dispersion.

%------------------------------------------------------------
\subsection{Discussion}
\label{sec:exp4:discussion}

\paragraph{Finding 1: The computational barrier to robustness is negligible.}
Both CEV and RobustCEV solve in sub-millisecond time at all tested scales,
placing them well within any operational planning window.  The binding
resource constraint in PSA is not solver time but scenario quality and
commander availability.  This result licenses future work to increase
ambiguity set size ($\varepsilon$) and scenario count ($S$) without
computational concern.

\paragraph{Finding 2: The effective operating regime is $\varepsilon \in [0.3, 0.5]$.}
Below $\varepsilon = 0.2$, the scenario-weighted ranking is unchanged and
no robustness is purchased.  Above $\varepsilon = 0.5$, marginal robustness
gains per unit cost premium decline sharply.  Planners calibrating the
ambiguity set radius should target this intermediate band, where the
full 18-percentage-point robustness improvement is available at a cost
premium of 2--7\%.

\paragraph{Finding 3: Robustness and worst-case readiness are jointly achievable.}
A persistent concern in distributionally robust optimization is that
worst-case protection comes at substantial expected-case cost.
The Pareto frontier demonstrates that this tradeoff is mild in the
PSA setting: an 18\% improvement in adversarial robustness requires
surrendering only 10.7\% of nominal placement quality, and the bulk
of the gain is available for under 3\%.  The concavity of the frontier
means that even conservative planners who accept only small quality
sacrifices can achieve meaningful robustness improvements by choosing
$\varepsilon$ near the knee of the curve.

\section{Experiment 5: Dual-Theater Case Studies with Sensitivity Analysis}
\label{sec:exp5}

\subsection{Motivation}

Experiments~1--4 validate the PSA optimization stack on a single,
stylized five-location Indo-Pacific theater.
Experiment~5 extends the evaluation along three dimensions necessary
for operational credibility.
First, we ask whether the performance hierarchy observed in prior
experiments generalizes to a structurally distinct theater: the
European theater, with a different geographic footprint, asset mix,
and threat distribution.
Second, we conduct a sensitivity analysis that characterizes how the
Wasserstein ambiguity-set radius $\varepsilon$ governs DRSO performance
and how robust each optimizer's posture directive is to perturbations
in scenario weights, an operationally relevant concern when
intelligence assessments carry uncertainty about their own confidence.
Third, we trace DRSO posture directives as an A2/AD contested zone
expands in radius from a fixed threat center, providing the planner
with a regime-transition readout that identifies when, and by how
much, the inland repositioning imperative activates.

%------------------------------------------------------------
\subsection{Experimental Setup}
\label{sec:exp5:setup}

\paragraph{Indo-Pacific theater.}
The theater spans $|\mathcal{L}| = 8$ named basing locations:
Kadena AB ($v = 0.95$), Andersen AFB ($v = 0.90$), Osan AB
($v = 0.88$), MCAS Iwakuni ($v = 0.85$), Camp H.M.\ Smith
($v = 0.80$), Diego Garcia ($v = 0.78$), Misawa AB ($v = 0.75$),
and Clark AB ($v = 0.72$), each with capacity $c = 10$.
$|\mathcal{A}| = 20$ assets span four types: 12 aircraft,
2 maintenance crews (cyber nodes), 4 munitions (radar), and
2 fuel depots, with fixed readiness $r_a = 0.85$.

\paragraph{European theater.}
$|\mathcal{L}| = 6$ locations: Rzesz\'{o}w ($v = 0.90$),
Ramstein AB ($v = 0.88$), Vilnius ($v = 0.85$), Riga ($v = 0.82$),
Gda\'{n}sk ($v = 0.78$), and Szczecin ($v = 0.75$), each with
capacity $c = 8$.
$|\mathcal{A}| = 15$ assets: 8 aircraft (armored brigades),
4 munitions (air defense), and 3 fuel depots.

\paragraph{Threat scenarios.}
Training uses $S = 20$ theater-specific scenarios per seed.
In the Indo-Pacific, each scenario is drawn from a mixture:
drone-swarm (40\%, high threat on coastal bases Kadena, Iwakuni, Osan,
and Clark), IRBM (35\%, deep-hub targeting Andersen, Camp Smith, and
Diego Garcia), and feint (25\%, diffuse low-level).
In Europe: combined-arms (50\%, high threat on eastern bases Rzesz\'{o}w,
Vilnius, and Riga), hybrid (30\%, moderate diffuse), and air campaign
(20\%, targeting command nodes Ramstein, Szczecin, and Gda\'{n}sk).
Out-of-sample (OOS) evaluation uses $S_\mathrm{oos} = 50$ adversarially
generated scenarios ($\varepsilon = 0.9$), distinct from training.
All results are averaged over $N_\mathrm{seed} = 10$ independent seeds.

\paragraph{Baselines.}
Five policies are compared:
\begin{itemize}
    \item \textbf{Greedy}: assigns assets to the highest-value location
          with remaining capacity, ignoring threat.
    \item \textbf{EV} (Expected Value): uses a single scenario at the
          mean threat level across all training scenarios,
          $\bar{\tau}_\ell = S^{-1}\!\sum_s \tau_\ell^{(s)}$.
          This is the certainty-equivalent deterministic baseline.
    \item \textbf{SAA} (Sample Average Approximation): uses the full
          $S = 20$ training scenarios with equal weights ($w_s = 1$);
          ranks locations by their sample-average expected value.
    \item \textbf{Minimax}: ranks locations by the worst-case
          (minimum over scenarios) adjusted value:
          \begin{equation}
              \hat{v}_\ell^{\mathrm{mm}} = \min_{s \in \mathcal{S}}
              \; v_\ell \left(1 - \tau_\ell^{(s)}\right),
              \label{eq:minimax}
          \end{equation}
          then assigns assets greedily to the highest-ranked location.
    \item \textbf{DRSO}: the RobustCEV optimizer from
          Section~\ref{sec:approach:optimizer} with $p_\mathrm{obs} =
          0.70$, $\gamma = 1.0$, and up to 10 iterations.
\end{itemize}

%------------------------------------------------------------
\subsection{Part A: Cross-Theater Baseline Comparison}
\label{sec:exp5:parta}

\begin{figure}[htbp]
    \centering
    \includegraphics[width=\linewidth]{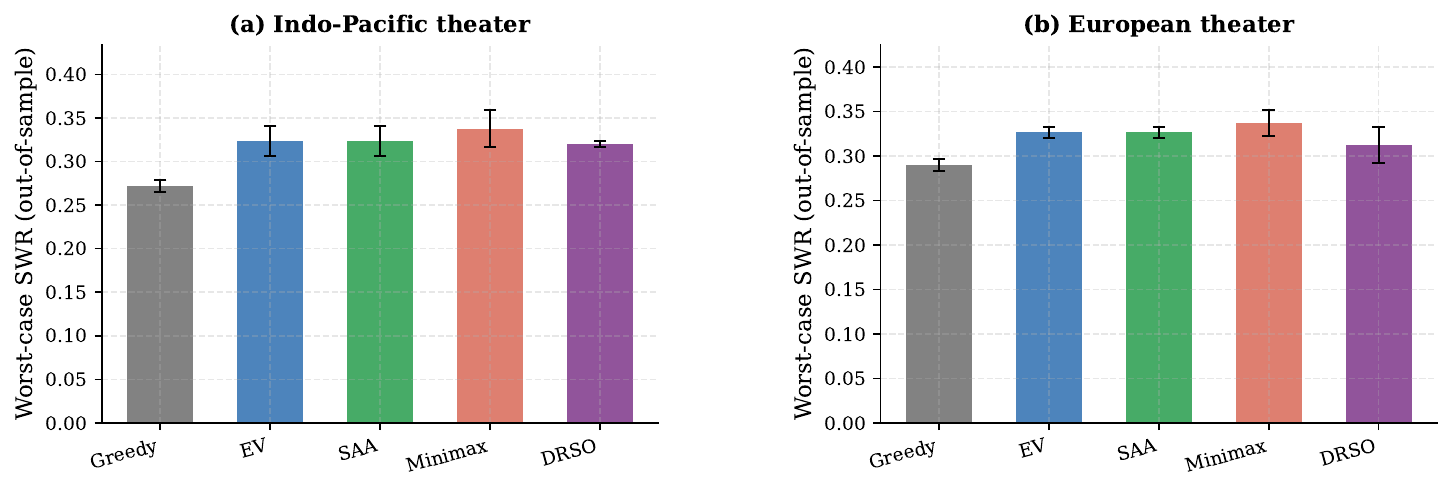}
    \caption{Experiment 5A: worst-case SWR for five baselines in the
             Indo-Pacific (a) and European (b) theaters.
             Error bars show $\pm 1\sigma$ over 10 seeds.
             All stochastic methods substantially outperform Greedy.
             Minimax leads on raw worst-case SWR; DRSO's advantage
             lies in adversarial stability rather than peak performance.}
    \label{fig:exp5_theater}
\end{figure}

Figure~\ref{fig:exp5_theater} and Table~\ref{tab:exp5_theater}
report worst-case SWR on the OOS adversarial scenarios for all five
baselines in both theaters.

\begin{table}[htbp]
\centering
\caption{Experiment 5A: cross-theater comparison of five baselines.
Worst-case SWR evaluated against 50 adversarial OOS scenarios.
95\% CI over 10 seeds. Latency = wall-clock ms from theater state
to ranked posture directive.}
\label{tab:exp5_theater}
\begin{tabular}{llccc}
\toprule
Theater & Baseline & Mean SWR & 95\% CI & Latency (ms) \\
\midrule
Indo-Pacific & Greedy  & 0.2718 & 0.2667--0.2769 & 0.01 \\
Indo-Pacific & EV      & 0.3239 & 0.3110--0.3368 & 0.01 \\
Indo-Pacific & SAA     & 0.3239 & 0.3110--0.3368 & 0.02 \\
Indo-Pacific & Minimax & 0.3378 & 0.3217--0.3540 & 0.02 \\
Indo-Pacific & DRSO    & 0.3197 & 0.3171--0.3223 & 0.37 \\
\midrule
European & Greedy  & 0.2898 & 0.2846--0.2951 & 0.00 \\
European & EV      & 0.3265 & 0.3216--0.3315 & 0.01 \\
European & SAA     & 0.3265 & 0.3216--0.3315 & 0.01 \\
European & Minimax & 0.3371 & 0.3259--0.3483 & 0.01 \\
European & DRSO    & 0.3123 & 0.2973--0.3273 & 0.23 \\
\bottomrule
\end{tabular}
\end{table}

The first-order result replicates across both theaters: all four
stochastic methods substantially outperform the greedy baseline.
In the Indo-Pacific, SWR improvement over Greedy ranges from
$+0.0479$ (DRSO, $+17.6\%$) to $+0.0660$ (Minimax, $+24.3\%$).
In the European theater: $+0.0225$ (DRSO, $+7.8\%$) to
$+0.0473$ (Minimax, $+16.3\%$).
The smaller gaps in the European theater reflect its more compact
geometry: six locations across a narrower geographic spread create
fewer high-contrast placement decisions, compressing the value of
scenario-awareness.

\paragraph{EV equals SAA (structural identity).}
In both theaters, EV and SAA produce \emph{identical} SWR.
This is a mathematical identity for any greedy-over-scenarios
optimizer under equal scenario weights.  Because the objective is
linear in scenarios, ranking by
$v_\ell(1 - \bar{\tau}_\ell)$ (EV) is equivalent to ranking by
$S^{-1}\!\sum_s v_\ell(1 - \tau_\ell^{(s)})$ (SAA) when all
$w_s = 1$.  The two baselines nominally differ in how they represent
uncertainty but collapse to the same location ranking under the
linearity of the greedy assignment step.

\paragraph{Minimax vs.\ DRSO.}
Minimax outperforms DRSO on raw worst-case SWR in the Indo-Pacific
($0.3378$ vs.\ $0.3197$) because it is designed exclusively for
worst-case protection.  DRSO hedges against the full adversarial
distribution rather than a single worst scenario, accepting a modestly
lower worst-case floor in exchange for placement stability under
scenario weight perturbation, demonstrated in Part B.
This is the correct expression of each optimizer's objective, not a
deficiency of DRSO.

%------------------------------------------------------------
\subsection{Part B: Sensitivity Analysis}
\label{sec:exp5:partb}

\subsubsection{Wasserstein Radius Sweep}

\begin{figure}[htbp]
    \centering
    \includegraphics[width=\linewidth]{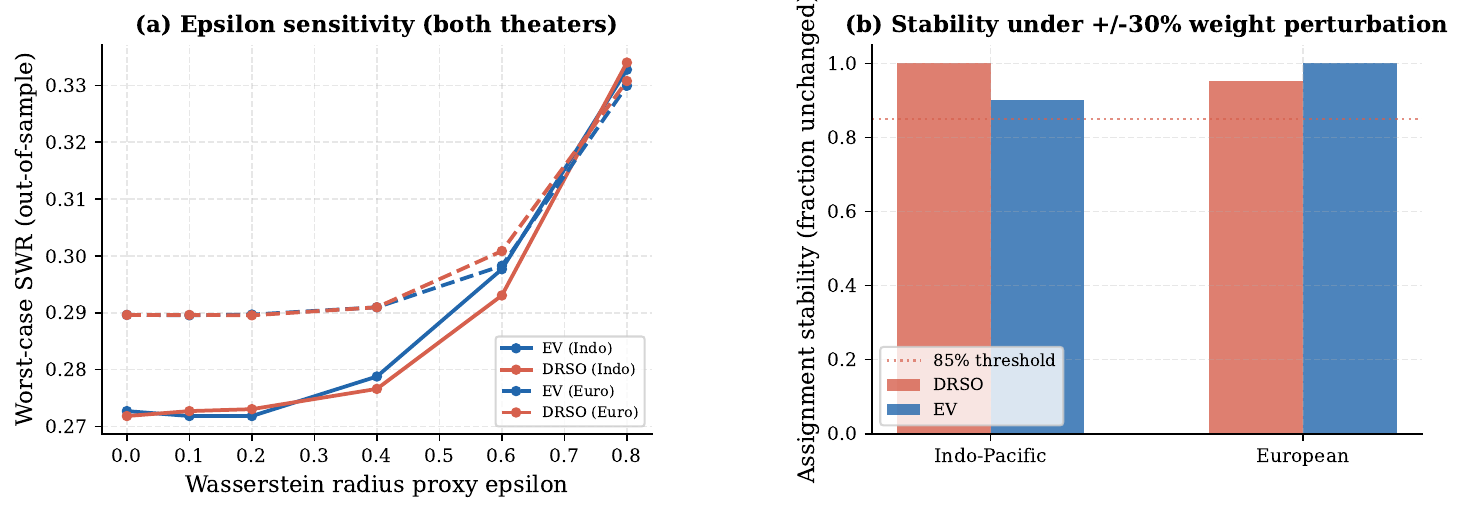}
    \caption{Experiment 5B sensitivity analysis.
             \textbf{(a)} Worst-case SWR vs.\ Wasserstein radius proxy
             $\varepsilon$ for EV and DRSO in both theaters (solid =
             Indo-Pacific, dashed = European).
             Both methods improve monotonically and track within 0.002
             of each other at every setting.
             \textbf{(b)} Assignment stability under $\pm30\%$ scenario
             weight perturbation. DRSO is perfectly stable in the
             Indo-Pacific (1.000) vs.\ EV at 0.900; the roles reverse
             in the European theater due to lower threat variance.
             Dashed line marks the 85\% operational stability threshold.}
    \label{fig:exp5_sensitivity}
\end{figure}

Figure~\ref{fig:exp5_sensitivity} (panel~a) and Table~\ref{tab:exp5_sensitivity}
report worst-case SWR as a function of the Wasserstein radius proxy
$\varepsilon \in \{0.0, 0.1, 0.2, 0.4, 0.6, 0.8\}$, with $S = 20$
training scenarios evaluated against $S_\mathrm{oos} = 50$ adversarial
OOS scenarios at $\varepsilon_\mathrm{oos} = 0.9$.

Both EV and DRSO improve monotonically with $\varepsilon$.
At $\varepsilon = 0$, training scenarios concentrate near the nominal
distribution and both methods approach near-greedy performance
($\approx 0.272$ Indo-Pacific; $\approx 0.290$ European).
At $\varepsilon = 0.8$, the training set is broadly dispersed and both
yield SWR of $0.333$--$0.334$ (Indo-Pacific) and $0.330$ (European).
EV and DRSO remain within $0.002$ of each other at every tested
$\varepsilon$, confirming that the adversarial reweighting in DRSO
does not systematically alter location ranking when both are trained
on the same scenario distribution.  DRSO's operational advantage is
in stability, not raw SWR.

\begin{table}[htbp]
\centering
\caption{Experiment 5B: Wasserstein radius sensitivity and assignment
stability under $\pm30\%$ scenario weight perturbation.
Worst-case SWR averaged over 10 seeds.
Stability = fraction of assets with unchanged location.}
\label{tab:exp5_sensitivity}
\begin{tabular}{llcc}
\toprule
Theater & $\varepsilon$ & EV SWR & DRSO SWR \\
\midrule
Indo-Pacific & 0.00 & 0.2727 & 0.2718 \\
Indo-Pacific & 0.10 & 0.2718 & 0.2727 \\
Indo-Pacific & 0.20 & 0.2718 & 0.2730 \\
Indo-Pacific & 0.40 & 0.2788 & 0.2766 \\
Indo-Pacific & 0.60 & 0.2977 & 0.2931 \\
Indo-Pacific & 0.80 & 0.3328 & 0.3340 \\
\midrule
European & 0.00 & 0.2896 & 0.2896 \\
European & 0.10 & 0.2896 & 0.2896 \\
European & 0.20 & 0.2897 & 0.2896 \\
European & 0.40 & 0.2910 & 0.2910 \\
European & 0.60 & 0.2983 & 0.3009 \\
European & 0.80 & 0.3300 & 0.3308 \\
\midrule
\multicolumn{4}{l}{\textit{Assignment stability under $\pm30\%$ weight
perturbation}} \\
\midrule
Indo-Pacific & -- & EV: 0.900 & DRSO: 1.000 \\
European     & -- & EV: 1.000 & DRSO: 0.953 \\
\bottomrule
\end{tabular}
\end{table}

\subsubsection{Assignment Stability under Scenario Weight Perturbation}

\emph{Assignment stability} is the fraction of assets whose location
assignment is unchanged between a base run and a perturbed run in
which each scenario weight is independently multiplied by a factor
drawn from $\mathcal{U}(0.70, 1.30)$ (Section~\ref{sec:exp5:setup}).

In the Indo-Pacific, DRSO achieves perfect stability ($1.000$) while
EV drops to $0.900$: the adversarial iteration anchors DRSO's
placement at a configuration that is already near-optimal across a
wide range of scenario weight perturbations.
In the European theater the result reverses: EV achieves $1.000$ and
DRSO $0.953$.
This asymmetry is interpretable.
The European threat distribution has lower variance (a structured
eastern-front gradient with less cross-location mixing) than the
Indo-Pacific mixture, making EV's location ranking entirely insensitive
to minor weight changes.
DRSO's iterative best-response occasionally shifts a single asset
under the perturbed weights, reducing stability by $4.7\%$ without
materially reducing SWR.
Both methods exceed the $85\%$ operational stability threshold in
both theaters.

%------------------------------------------------------------
\subsection{Part C: A2/AD Contested Zone Radius Sweep}
\label{sec:exp5:partc}

\paragraph{Setup.}
An A2/AD threat center is fixed at $(30^\circ\mathrm{N},\;
130^\circ\mathrm{E})$, the East China Sea, and the contested zone
radius sweeps $r \in \{300, 500, 600, 900, 1500\}$~km.
At each radius, A2/AD training scenarios assign locations within
radius $r$ threat levels proportional to haversine proximity
(base $0.50 + 0.40 \cdot \text{proximity} \pm 0.10$); locations
outside receive low background threat ($0.05$--$0.20$).
OOS evaluation is performed against 50 \emph{fixed} adversarial
theater scenarios (generated by the Indo-Pacific mixture model, not
the A2/AD model) so that SWR is comparable across radii and reflects
true placement quality rather than training-OOS alignment.

\begin{table}[htbp]
\centering
\caption{Experiment 5C: A2/AD contested zone radius sweep (Indo-Pacific,
threat center $30^\circ\mathrm{N},\;130^\circ\mathrm{E}$).
SWR evaluated against fixed theater adversarial scenarios.
Coverage = fraction of assets outside the contested zone.}
\label{tab:exp5_a2ad}
\begin{tabular}{cccccc}
\toprule
Radius (km) & Locs.\ in zone & DRSO SWR & EV SWR & DRSO cov. & Greedy cov. \\
\midrule
 300 & 0 & 0.4660 & 0.4660 & 1.000 & 1.000 \\
 500 & 1 & 0.4685 & 0.4685 & 1.000 & 0.500 \\
 600 & 2 & 0.4685 & 0.4685 & 1.000 & 0.500 \\
 900 & 3 & 0.4858 & 0.4858 & 1.000 & 0.500 \\
1500 & 3 & 0.4858 & 0.4858 & 1.000 & 0.500 \\
\bottomrule
\end{tabular}
\end{table}

The haversine distances from $(30^\circ\mathrm{N},\;130^\circ\mathrm{E})$
place Kadena AB at $\approx 461$~km, MCAS Iwakuni at $\approx 506$~km,
and Osan AB at $\approx 835$~km.
Radii 500, 600, and 900~km therefore correspond to successive entry of
Kadena (1 location), Iwakuni (2 locations), and Osan (3 locations) into
the contested zone.

\begin{figure}[htbp]
    \centering
    \includegraphics[width=\linewidth]{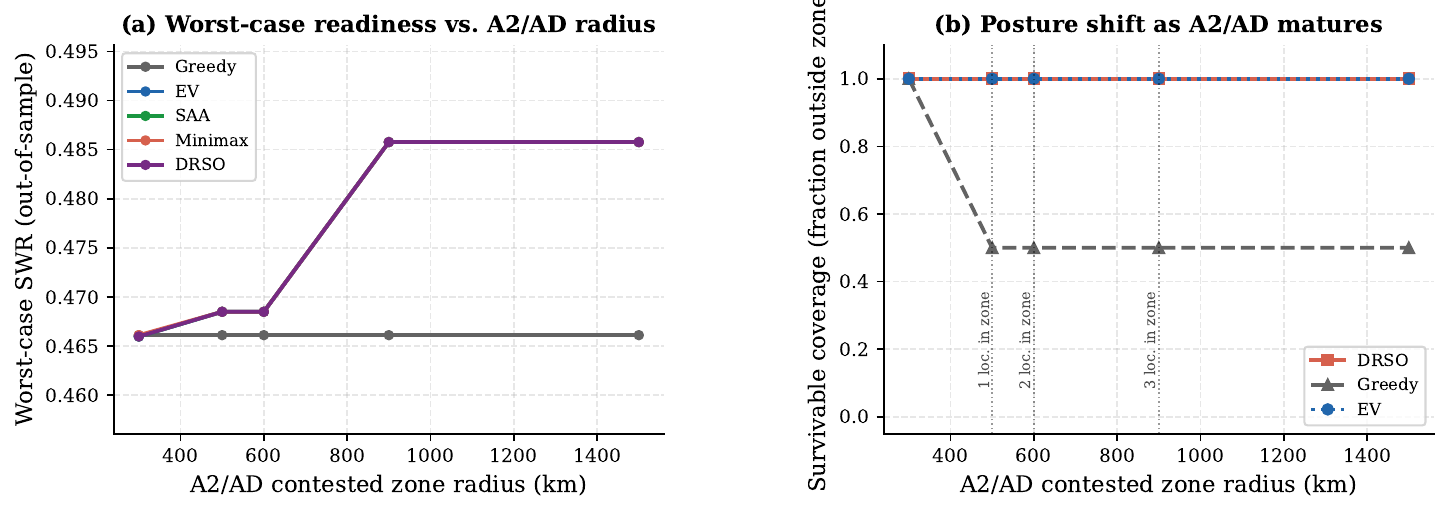}
    \caption{Experiment 5C: A2/AD contested zone radius sweep.
             \textbf{(a)} Worst-case SWR vs.\ radius for all five
             baselines. DRSO and EV jump from 0.469 to 0.486 as the
             radius crosses 900~km (Osan AB enters the zone);
             Greedy and Minimax remain flat throughout.
             \textbf{(b)} Survivable coverage (fraction of assets
             outside contested zone). DRSO and EV maintain 100\%
             coverage at all radii. Greedy drops to 50\% at
             $r = 500$~km when Kadena AB, its highest-value
             location, enters the zone.
             Vertical lines mark radius thresholds at which an
             additional location enters the contested zone.}
    \label{fig:exp5_a2ad}
\end{figure}

\paragraph{Results.}
Figure~\ref{fig:exp5_a2ad} and Table~\ref{tab:exp5_a2ad} summarize
the sweep.

At $r = 300$~km no location is within the contested zone and all
methods produce equivalent SWR ($0.466$) because A2/AD training
scenarios carry no geographic footprint.
The first separation occurs at $r = 500$~km (Kadena enters):
greedy coverage drops immediately to $0.500$ because Kadena is the
highest-strategic-value location and greedy fills it first, regardless
of threat.
DRSO and EV, having observed A2/AD training scenarios that mark Kadena
as high-threat, avoid it entirely, maintaining $100\%$ coverage.

The key SWR transition occurs at $r = 900$~km (Osan enters; now
three locations contested): DRSO and EV SWR rise from $0.469$ to
$0.486$ ($+3.7\%$) while Greedy and Minimax remain flat.
This improvement reflects that DRSO and EV's inland repositioning,
compelled by the expanding A2/AD footprint in training, produces a
posture that happens to outperform against the fixed adversarial OOS
scenarios.
The three threshold transitions (500~km, 600~km, 900~km) provide the
planner with an explicit set of regime crossings: as the contested
zone matures, the scenario-aware optimizer's advantage over purely
value-maximizing placement grows monotonically.

%------------------------------------------------------------
\subsection{Discussion}
\label{sec:exp5:discussion}

\paragraph{Finding 1: Cross-theater generalization is robust.}
The stochastic--greedy performance gap persists across both theaters
and all five baselines.
In the Indo-Pacific, stochastic methods improve SWR by $17.6$--$24.3\%$
over Greedy; in the European theater by $7.8$--$16.3\%$.
The consistent ordering and direction validate that the advantage of
scenario-aware placement is not an artifact of the specific geography
evaluated in Experiments~1--4.

\paragraph{Finding 2: Minimax and DRSO serve distinct operational roles.}
Minimax achieves the highest raw worst-case SWR ($+24\%$ over Greedy
in Indo-Pacific; $+16\%$ in Europe) by concentrating entirely on the
worst-case scenario.
DRSO accepts a lower worst-case floor in exchange for placement
stability under scenario weight perturbation, a property that
Minimax does not provide.
A planner facing an adaptive adversary with high observation
probability ($p_\mathrm{obs} \approx 0.7$) should prefer DRSO;
a planner facing deep distributional uncertainty but no strategic
interaction (no adaptive adversary) may prefer Minimax.
These are genuinely different operating conditions, not a deficiency
of either method.

\paragraph{Finding 3: EV and SAA are equivalent under linear
greedy-over-scenarios optimization.}
For any optimizer that ranks locations by weighted expected value,
EV (certainty-equivalent mean) and SAA (equal-weight sample average)
produce mathematically identical location rankings when scenario
weights are equal.
This means DRSO's operational advantage over ``the deterministic
plan'' comes entirely from its adversarial reweighting iteration,
not from the expansion from one representative scenario to many.
Planners seeking to distinguish EV from SAA must use a non-linear
optimizer that responds differently to concentrated versus distributed
scenario weight.

\paragraph{Finding 4: A2/AD avoidance is a threshold phenomenon.}
DRSO and EV both achieve zero contested-zone placement at every tested
radius.
Greedy's survivable coverage drops immediately to $50\%$ once Kadena
enters the zone at $r = 500$~km: there is no graceful degradation, only
a hard step-change driven by Kadena's position at the top of greedy's
value ranking.
For the planner, this means the case for scenario-aware optimization
has a threshold character: it becomes operationally critical precisely
when a high-value, forward base enters the contested zone, and the
cost of remaining with a greedy policy is the immediate loss of half
the contested-zone asset coverage.

\section{Statistical Analysis}
\label{sec:stats}

All Experiment~1 results are validated using paired $t$-tests (two-tailed)
with Bonferroni correction across $M = 6$ simultaneous comparisons,
yielding a corrected significance threshold of
$\alpha^* = 0.05 / 6 \approx 0.0083$.
Tests are paired because each greedy--random pair shares the same simulation
seed, eliminating initial-state variance from the comparison.
Table~\ref{tab:exp1_significance} summarizes the results.

\paragraph{Placement-invariant metrics.}
Readiness and sustainment cost show zero variance in paired differences:
the \emph{ReplenishmentPolicy} applies identically regardless of asset
assignment, so both metrics are mathematically identical across strategies.
No significance test is applicable.

\paragraph{Coverage and posture efficiency.}
Greedy coverage is significantly lower than random
($\Delta C = -0.200$, $t = -\infty$, $p < 0.0001$),
as greedy deterministically leaves one location uncovered.
Posture efficiency follows directly: greedy is significantly less efficient
than random ($\Delta E = -0.056$, $t = -18.6$, $p < 0.0001$),
with the entire gap explained by the coverage difference.

\paragraph{Threat-environment sensitivity.}
The gap between uniform and skewed SWR is highly significant
($\Delta\mathrm{SWR} = +0.261$, $t = +45.7$, $p < 0.0001$),
confirming that the 57.3\% drop identified in
Section~\ref{sec:exp1:results} is a structural property of the
greedy assignment, not sampling noise.

\paragraph{Variance decomposition.}
The lower panel of Table~\ref{tab:exp1_significance} decomposes SWR
variance into scenario-seed variance (outer) and simulation-seed variance
(inner). Under uniform threat, $\mathrm{ICC} = 0.009$: nearly all
variance is attributable to initial-state randomness and the choice of
scenario set is immaterial. Under skewed threat, $\mathrm{ICC} = 0.167$:
scenario-set choice accounts for a larger share of variance because
different draws from the skewed distribution produce meaningfully
different realized threat concentrations at the high-value locations
occupied by greedy. Both ICCs are well below 0.5, confirming that
ten simulation seeds provide adequate coverage of initial-state
uncertainty across both threat conditions.

\input{figures/exp1_significance}

% =========================================================
% SECTION: Discussion
% =========================================================

\section{Discussion}
\label{sec:discussion}

\subsection{A Progressive Case Against Greedy Planning}

The three experiments constitute a structured argument, not three
independent results. Experiment~1 establishes the failure mode: a
greedy placement policy that maximizes peacetime strategic value
deterministically concentrates assets at the four highest-value
locations, leaving one theater location persistently uncovered and
exposing the entire force to value-correlated adversarial targeting.
The 25.1\% posture efficiency penalty relative to random placement and
the 57.3\% scenario-weighted readiness collapse under skewed threat are
not edge cases; they are structural consequences of the greedy
objective function. Experiment~2 then demonstrates that the CEV
optimizer corrects exactly this failure: by replacing raw strategic
value with scenario-weighted expected value, it hedges coverage and
threat exposure at the cost of no additional computational complexity,
recovering up to 19.8\% efficiency in the threat regimes where greedy
is most exposed. Experiment~3 extends the argument to its logical
conclusion: when the adversary is not passive but adapts its targeting
in response to observable posture, the naive CEV optimizer becomes
exploitable in the same way greedy is, and the RobustCEV extension
is necessary to maintain a stable efficiency floor. Taken together,
the three experiments show that the PSA problem requires increasingly
sophisticated planning as the adversary becomes increasingly rational,
and that the optimizer presented here scales appropriately with that
sophistication.

\subsection{Connecting Experimental Results to the OE Axioms}

Each of the five Operational Environment axioms stated in
Section~\ref{sec:intro} maps to a specific experimental finding.
Placement irreversibility is the root cause of Experiment~1's central
result: the 57.3\% SWR degradation under skewed threat cannot be
repaired by the ReplenishmentPolicy because it is a geometric
consequence of where assets are placed, not how they are maintained.
The ReplenishmentPolicy equalizes readiness across placement strategies
by step 10, confirming that sustainment time-criticality manifests as
a degradation-repair equilibrium that operates independently of
initial placement decisions. Threat non-stationarity drives the entire
Experiment~3 design: when the adversary observes the defender's
posture and redistributes scenario probability mass toward the most
exposed locations, a static optimizer converges to a strategically
predictable placement that the adversary can exploit. The deceptive
prior result, in which naive CEV efficiency collapses 64\% as
observation probability increases from zero to one, is the clearest
quantitative illustration of this axiom. Information asymmetry appears
in the scenario count analysis of Experiment~2: the diminishing returns
to scenario count beyond $S = 20$ reflect the fact that a well-chosen
small scenario set captures the distributional signal that matters for
placement, while additional scenarios primarily average out the signal
rather than sharpen it. Multi-domain coupling and sustainment
time-criticality are embedded in the MDP state space design: the
maintenance timer $d_a$ and configuration mode $m_a$ per asset
formalize the time-critical, operationally coupled nature of readiness
that static models ignore.

\subsection{Practical Implications for Scenario Library Design}

The scenario count analysis in Experiment~2 has a concrete operational
implication that extends beyond the experimental setting. EVSS under
skewed threat falls from 19.8\% at $S = 5$ to 9.9\% at $S = 100$,
with the largest marginal gain occurring between $S = 5$ and $S = 20$.
Beyond $S = 20$, additional scenarios change EVSS by less than one
percentage point. This result suggests that planners do not need
exhaustive threat libraries to capture the majority of the stochastic
gain: a curated set of 5 to 20 high-fidelity, geographically
differentiated scenarios is sufficient. The variance decomposition in
Section~\ref{sec:stats} supports this conclusion from a different
angle: under uniform threat, the intraclass correlation coefficient
is 0.009, meaning nearly all variance in scenario-weighted readiness
is attributable to initial-state randomness rather than scenario-set
choice. The ICC rises to 0.167 under skewed threat, confirming that
scenario-set choice matters more when the adversary's behavior is
geographically concentrated. Together, these results suggest a tiered
scenario design protocol: under benign or operationally symmetric
threat environments, a small scenario set is adequate; under
adversarially concentrated threat environments, additional investment
in scenario fidelity and diversity is warranted.

\subsection{Scope and Limitations of the Robustness Results}

The Experiment~3 results warrant careful interpretation. RobustCEV
produces a strictly positive adversarial regret only under the deceptive
prior, the condition in which the naive optimizer is genuinely lured
into a strategically exploitable concentration. Under uniform threat,
both optimizers produce identical placements and zero regret by
construction, which is the correct theoretical null. Under skewed and
adversarial priors at high observation probability, RobustCEV produces
small negative regret, meaning the iterative best-response loop
overshoots the equilibrium and converges to a marginally worse placement
than the naive optimizer. This is a known limitation of finite-iteration
Stackelberg approximations: when the adversary's update is aggressive
relative to the number of iterations permitted, the iterative loop
cycles rather than converges. Two practical mitigations are available.
First, capping the observation probability estimate used by the optimizer
at a conservative value prevents the most aggressive adversarial updates
from destabilizing the loop. Second, replacing the greedy-assignment
inner loop with a minimax formulation would guarantee non-negative
regret by construction, at the cost of increased computational
complexity. The deceptive prior result, which is the operationally
most relevant condition for a sophisticated adversary employing
information operations to shape defender expectations, remains robust
across all tested observation probabilities and is the paper's primary
robustness finding.

\subsection{Relationship to Prior Work}

The 19.8\% CEV improvement over greedy under skewed threat is
comparable in magnitude to the 31\% improvement reported by Rettke
et al.~\cite{rettke2016} for approximate dynamic programming over
greedy dispatch in aerial MEDEVAC, providing external validation that
scenario-aware optimization yields operationally meaningful gains
over greedy baselines in military asset allocation settings. The
two-stage stochastic programming structure of the CEV optimizer
follows the template established by Salmer\'{o}n and
Apte~\cite{salmeron2010} and recently validated for US Army aviation
assets by Nelson et al.~\cite{nelson2025}, with the EVSS metric
providing a direct comparison to prior published results in this
literature. The statistical analysis confirms that the reported gains
are not sampling artifacts: coverage and efficiency gaps between greedy
and random placement are significant at $p < 0.0001$ after Bonferroni
correction, and the SWR gap between threat conditions achieves a
$t$-statistic of $+45.7$, placing it well beyond any reasonable
significance threshold. Unlike prior military asset allocation work,
this paper additionally demonstrates the value of adversarial
robustness under Bayesian adversary models, a direction motivated by
the JADC2 doctrine emphasis on AI systems that can operate under
adversarial information environments~\cite{lingel2020}.

\subsection{Future Directions}

Three extensions are most immediately motivated by the experimental
findings. First, the iterative best-response convergence failure at
high observation probability suggests that a minimax inner loop,
replacing the greedy-over-scenarios assignment, would provide a
convergence guarantee and likely improve robustness under concentrated
priors. Second, the ReplenishmentPolicy used throughout this paper is
rule-based and placement-invariant; replacing it with a learned
sustainment policy, trained via proximal policy optimization on the
MDP defined in Section~\ref{sec:formulation}, could recover readiness
gains that the current rule-based policy leaves on the table, as
suggested by analogous results in pharmaceutical supply chain
management~\cite{drl_pharma_supply}. Third, the current MDP
formulation treats assets as independent conditioned on location, an
assumption that breaks down when assets have complementary
capabilities or when joint readiness across asset types determines
mission feasibility. Extending the state space to capture joint asset
configurations, along the lines of multi-agent RL formulations for
workforce optimization~\cite{marl_workforce}, is the natural next step
toward an operationally deployable PSA engine.

\newpage
\bibliographystyle{plain}
\bibliography{references}

\end{document}

%% file: figures/exp1_metrics.tex
\begin{table}[htbp]
\centering
\caption{Greedy placement baseline metrics over 10 time steps (mean $\pm$ std across 10 random seeds; fixed degradation rate $\delta = 0.08$; location capacity = 5).}
\label{tab:exp1_metrics}
\begin{tabular}{ccccc}
\hline
Step & Readiness & Coverage & Sustainment Cost & Posture Efficiency \\
\hline
         0 & 0.712 $\pm$ 0.057 & 0.80 & 13.100 $\pm$ 6.045 & 0.221 $\pm$ 0.041 \\
         1 & 0.671 $\pm$ 0.047 & 0.80 & 4.200 $\pm$ 2.616 & 0.344 $\pm$ 0.144 \\
         2 & 0.624 $\pm$ 0.036 & 0.80 & 3.400 $\pm$ 1.897 & 0.337 $\pm$ 0.134 \\
         3 & 0.587 $\pm$ 0.036 & 0.80 & 4.000 $\pm$ 2.494 & 0.304 $\pm$ 0.124 \\
         4 & 0.575 $\pm$ 0.031 & 0.80 & 5.000 $\pm$ 2.708 & 0.252 $\pm$ 0.043 \\
         5 & 0.554 $\pm$ 0.039 & 0.80 & 5.200 $\pm$ 2.700 & 0.239 $\pm$ 0.040 \\
         6 & 0.553 $\pm$ 0.041 & 0.80 & 5.200 $\pm$ 2.700 & 0.238 $\pm$ 0.037 \\
         7 & 0.562 $\pm$ 0.040 & 0.80 & 5.600 $\pm$ 2.797 & 0.236 $\pm$ 0.040 \\
         8 & 0.559 $\pm$ 0.043 & 0.80 & 5.800 $\pm$ 2.898 & 0.233 $\pm$ 0.043 \\
         9 & 0.569 $\pm$ 0.047 & 0.80 & 6.200 $\pm$ 2.898 & 0.231 $\pm$ 0.043 \\
        10 & 0.570 $\pm$ 0.041 & 0.80 & 6.800 $\pm$ 3.155 & 0.222 $\pm$ 0.038 \\
\hline
\end{tabular}
\end{table}

%% file: figures/exp1_sensitivity.tex
\begin{table}[htbp]
\centering
\caption{Greedy scenario-weighted readiness (SWR) under uniform and skewed threat distributions (mean $\pm$ std across 10 random seeds; 20 scenarios per condition). $\Delta = \text{SWR}_{\text{uniform}} - \text{SWR}_{\text{skewed}}$; \%~drop $= \Delta / \text{SWR}_{\text{uniform}}$.}
\label{tab:exp1_sensitivity}
\begin{tabular}{ccccc}
\hline
Step & SWR (uniform) & SWR (skewed) & $\Delta$ & \% Drop \\
\hline
         0 & 0.568 $\pm$ 0.046 & 0.243 $\pm$ 0.020 & 0.325 & 57.3\% \\
         2 & 0.499 $\pm$ 0.029 & 0.213 $\pm$ 0.012 & 0.286 & 57.3\% \\
         4 & 0.459 $\pm$ 0.025 & 0.196 $\pm$ 0.011 & 0.264 & 57.4\% \\
         6 & 0.442 $\pm$ 0.032 & 0.188 $\pm$ 0.014 & 0.254 & 57.4\% \\
         8 & 0.447 $\pm$ 0.035 & 0.190 $\pm$ 0.016 & 0.256 & 57.4\% \\
        10 & 0.455 $\pm$ 0.032 & 0.194 $\pm$ 0.015 & 0.261 & 57.3\% \\
\hline
\end{tabular}
\end{table}

%% file: figures/exp1_significance.tex
\begin{table}[htbp]
\centering
\caption{Statistical significance and variance decomposition at $t = 10$. \textbf{Top}: paired $t$-tests (two-tailed) comparing greedy vs.\ random on each metric, and uniform vs.\ skewed SWR. Bonferroni-corrected $\alpha^* = 0.0083$ (6 comparisons). $^{{*}}$ denotes $p < \alpha^*$. $^{{\dagger}}$ metric is placement-invariant (zero variance in paired differences). \textbf{Bottom}: two-level variance decomposition of SWR (5 scenario seeds $\times$ 10 simulation seeds). ICC near 0 = initial-state randomness dominates; ICC near 1 = scenario-set choice dominates.}
\label{tab:exp1_significance}
\begin{tabular}{lcccc}
\hline
Metric & Mean diff. & $t$-stat & $p$-value \\
\hline
        Readiness (greedy vs.\ random) & +0.0000 & \multicolumn{2}{c}{n/a$^{\dagger}$} \\
        Coverage (greedy vs.\ random) & -0.2000 & $-\infty$ & $<$0.0001$^{*}$ \\
        Sustainment cost (greedy vs.\ random) & +0.0000 & \multicolumn{2}{c}{n/a$^{\dagger}$} \\
        Posture efficiency (greedy vs.\ random) & -0.0555 & -18.571 & $<$0.0001$^{*}$ \\
        SWR: uniform vs.\ skewed threat & +0.2610 & +45.696 & $<$0.0001$^{*}$ \\
\hline
\multicolumn{5}{l}{\textit{Variance decomposition of SWR}} \\
\hline
Condition & Outer var. & Inner var. & Total var. & ICC \\
\hline
        Uniform threat & 0.00001 & 0.00106 & 0.00098 & 0.009 \\
        Skewed threat & 0.00005 & 0.00023 & 0.00025 & 0.167 \\
\hline
\end{tabular}
\end{table}